\documentclass[a4paper,fleqn]{cas-sc}

\usepackage[numbers]{natbib}
\usepackage{subcaption}
\usepackage{longtable}

\usepackage[section]{placeins}

\def\tsc#1{\csdef{#1}{\textsc{\lowercase{#1}}\xspace}}
\tsc{WGM}
\tsc{QE}
\tsc{EP}
\tsc{PMS}
\tsc{BEC}
\tsc{DE}

\usepackage{color}

\begin{document}

\let\WriteBookmarks\relax
\def\floatpagepagefraction{1}
\def\textpagefraction{.001}

\shorttitle{Understanding the Effects of Human-written Paraphrases in LLM-generated Text Detection}

\shortauthors{Lau \& Zubiaga}

\title [mode = title]{Understanding the Effects of Human-written Paraphrases in LLM-generated Text Detection}                      


%
\author{Hiu Ting Lau}[orcid=0009-0008-7085-1800]



\ead{ml21305@qmul.ac.uk}



\affiliation{organization=School of Electronic Engineering and Computer Science, Queen Mary University of London, London E1 4NS}

\author{Arkaitz Zubiaga}[orcid=0000-0003-4583-3623]

\ead{a.zubiaga@qmul.ac.uk}



\begin{abstract}
Natural Language Generation has been rapidly developing with the advent of large language models (LLMs). While their usage has sparked significant attention from the general public, it is important for readers to be aware when a piece of text is LLM-generated. This has brought about the need for building models that enable automated LLM-generated text detection, with the aim of mitigating potential negative outcomes of such content. Existing LLM-generated detectors show competitive performances in telling apart LLM-generated and human-written text, but this performance is likely to deteriorate when paraphrased texts are considered. In this study, we devise a new data collection strategy to collect Human \& LLM Paraphrase Collection (HLPC), a first-of-its-kind dataset that incorporates human-written texts and paraphrases, as well as LLM-generated texts and paraphrases. With the aim of understanding the effects of human-written paraphrases on the performance of state-of-the-art LLM-generated text detectors OpenAI RoBERTa and watermark detectors, we perform classification experiments that incorporate human-written paraphrases, watermarked and non-watermarked LLM-generated documents from GPT and OPT, and LLM-generated paraphrases from DIPPER and BART. The results show that the inclusion of human-written paraphrases has a significant impact of LLM-generated detector performance, promoting TPR@1\%FPR with a possible trade-off of AUROC and accuracy.
\end{abstract}



\maketitle

\section{Introduction}
\label{sec:introduction}

Large language models (LLMs) have become essential in Natural Language Processing (NLP) thanks to their advanced capabilities for text processing and generation, which is achieved through analysis of patterns and relationships between words and sentences using transformer models \cite{zubiaga2024natural}. Consequently, LLMs have had a significant impact on Natural Language Generation (NLG), as they have provided improved capacity for automatically generating high quality text \cite{barreto2023generative}.

While the advancement of LLMs in the context of NLG has aided tasks such as machine translation \citep{zhang2023prompting} and text summarization \citep{zhang2024benchmarking}, it has also given rise to undesired social problems, including intentional malicious usage, ethical concerns and information inaccuracy. This has brought about the need for researching the development of methods for automated LLM-generated text detection which distinguishes if a text is human- or LLM-generated \cite{frohling2021feature}. Currently, there are 2 major streams of LLM-generated text detectors: (i) zero-shot classifiers \cite{2,1}, which identify LLM-generated text based on the pattern and characteristics of the input, and (ii) watermark detectors \cite{3}, which rely on detecting the presence of watermarks which are imprinted into the text during the generation process \cite{kirchenbauer2023watermark}, and are effective in the cases where the watermarks have been added by the LLM. The detectors then examine the input, classifying it as LLM-generated if the level of watermarking exceeds a set threshold, or as human-generated otherwise.

Both kinds of detectors have demonstrated excellent performance in LLM-generated text detection. However, research testing these detectors has primarily focused on datasets involving texts which are exclusively generated by humans or by LLMs. There can be, however, more complicated cases, such as paraphrased texts, which have been seldom considered in previous research. Paraphrasing is defined as the rewriting of context in a simpler and shorter form \cite{4}; an LLM-generated text which is then paraphrased by humans leads to modified texts where the statistical properties of watermarks in the LLM-generated text is no longer identifiable. Since the above detectors perform classification based on token patterns and watermarks, paraphrasing could effectively evade both zero-shot classifiers and watermark detectors while preserving semantic information from the original LLM-generated text. It is important to identify that a text originated from an LLM, despite having been subsequently paraphrased, as this can still be leveraged for massive generation of texts for malicious purposes.

In this work, we are the first to comprehensively study the effectiveness of LLM-generated text detectors in the presence of human-paraphrased texts, in turn assessing the impact of these edited texts on the model performance. In this study, we aim to address this problem by using human-written paraphrases for classification, with the notion that human-written paraphrases and LLM-generated paraphrases might contain different characteristics, which potentially improve the classifiers’ performances. We set forth the following research question: ``What are the effects of including human-written paraphrases in LLM-generated text detection?'' With this aim and research question in mind, we make the following contributions:

\begin{itemize}
 \item We perform a review of the literature to investigate the existing NLG developments, the importance of LLM-generated text detection, and the performances of existing detectors.
 \item We describe a data collection process which enables us to build and release the Human \& LLM Paraphrase Collection (HLPC) dataset with human-generated and LLM-generated documents, along with their paraphrases.
 \item We perform classification experiments using state-of-the-art AI paraphrasers and detectors, along with human-written paraphrases.
 \item Our study contributes to the domain of LLM-generated text detection by examining the effects of including human-written paraphrases in classification and providing insights on data inclusion in future detector building.
\end{itemize}

\section{Related Work}
\label{sec:related-work}

We review existing research in NLG with a focus on LLMs, the importance of LLM-generated text detection and existing detection models, following with a discussion of the main research gaps that our study addresses.

\subsection{Natural Language Generation (NLG)}
\label{ssec:nlg}

NLG represents a substantial branch of research within NLP, where existing NLG tasks include question-answering \cite{allam2012question}, text summarization \cite{el2021automatic}, and machine translation \cite{lopez2008statistical}.

\paragraph{\textbf{LLMs in NLG.}}
Very recently, the development of LLMs has brought significant improvements to NLG, primarily due to their ability of learning linguistic patterns from very large-scale corpora. Before the adoption of LLMs, two techniques were used for NLG. The earliest NLG systems used templates and rules \cite{6} and the later systems utilized conditional probability between words to account for context dependency \cite{7}. Fundamentally, these systems lack flexibility and adaptability since text generation is restricted, resulting in unfavorable generation patterns, such as inaccuracy in question answering in the rule-based model and word repetitions in the probabilistic model.

Recent research proposed different LLMs that adopt deep learning and neural networks in NLP, contributing to a remarkable improvement in NLG \cite{7}. With the use of a transformer architecture, the models can capture long-range text dependencies with positional encoding, allowing the models to understand both in-words and in-sentence relationships. The models can also be fine-tuned to cater to specific NLG needs. Coupled with large-scale datasets and parameters, LLMs can understand complex linguistic patterns and relationships. For example, GPT-3 is trained with 499 billion tokens and 175 billion parameters \cite{8,9}. This promotes models’ learning of language representations, including syntactic, contextual and semantic information \cite{7}. Text data from large corpora are used to train LLMs, such as CommonCrawl, WebText2, BookCorpus, etc \cite{8,9}, which allow the models to learn knowledge from a broad range of disciplines. With the above advancements, LLMs can thus generate human-like outputs solely using user prompts as inputs \cite{10}. As a result, the use of LLMs along with a simple chat interface has attracted massive usage and attention from the general public \cite{11}, with over 180 million users for OpenAI ChatGPT \cite{12}.

While LLMs provide a new direction for text generation in NLG, they are also being widely used to support evaluation of NLG outputs \cite{li2024leveraging}, which is however beyond the scope of this study.

\paragraph{\textbf{Paraphrase Generation.}}
One of the NLG tasks for which LLMs have brought a significant boost is paraphrase generation, with various AI paraphrasers built on top of existing LLMs. Paraphrasing is defined as the rewriting of context in a simpler and shorter form \cite{4}, and is used extensively to avoid content to be flagged as plagiarism. A paraphrase generator takes sentences or paragraphs as inputs, and creates rewritten outputs which preserve the semantics of the original text \cite{7}. Currently, there are 2 types of AI paraphrasers: (i) systems that are inherently built for paraphrasing, specifically built to paraphrase text automatically and evade LLM-generated text detectors; an example of this is DIPPER \cite{10}, which could paraphrase long paragraphs and control output diversity, and (ii) AI chatbots that receive paraphrasing prompts to produces paraphrases; an example of this is T5-paraphraser Parrot \cite{13}.

To account for the quality of paraphrases, grammar accuracy and content semantic preservation are considered either with human experts or automatic model evaluation. In \cite{14}, human evaluation was conducted for the paraphrases generated from DIPPER and Llama-2-7B-Chat. Paraphrases from both models exhibit high ratings in terms of content preservation and grammar accuracy. \cite{10} uses the P-SP model to compute semantic similarity scores which reflects the level of contextual relationship between the paraphrases and original text. DIPPER effectively paraphrases text with high semantic similarity. This shows that existing AI paraphrasers, coupled with the use of LLM, have shown great advancement in aiding automatic paraphrase generation.

\subsection{Importance of LLM-generated Text Detection}
\label{ssec:importance-llm-detection}

Despite the convenience brought by LLMs for improved NLG, they can also cause various problems with a negative impact on society, which has motivated the need for investigating methods for LLM-generated text detection. Next, we discuss three key problems arising from the use of LLMs, especially when the generated texts is not flagged or labelled as being LLM-generated: intentional malicious usage, ethical concerns and information inaccuracy \cite{zhao2024silent}.

\paragraph{\textbf{Intentional malicious usage.}} Malicious actors may exploit LLMs to produce fake content or to produce content in circumstances where LLM-generated content is unacceptable. Examples include the creation of fake news in political elections to denigrate competitors \cite{15} and the creation of writing or code to then claim full credit for it in an academic environment \cite{3}. As proposed in \cite{1}, malicious actors can be categorized into 3 levels: low-skilled, moderate-skilled and advanced-skilled based on their programming level and available resources. Moderate-skilled actors could already produce fake news or build spambots for social media, let alone the adverse effects that advanced-skilled actors could bring.

\paragraph{\textbf{Ethical concerns.}} Ethical considerations concerning gender, race and ethnicity bias are raised attributing to the inherent sampling bias of the LLM, which potentially leads to social inequality and discrimination \cite{an2024large}. As presented in Section \ref{ssec:nlg}, LLMs are trained from very large corpora. However, the sampling of content from these corpora might itself be biased. For example, a study found that Reddit data used to train GPT-2 \cite{16} is composed of context generated mainly by young males in the United States \cite{17}. Since the training of models gives equal weighting for each document in the sample \cite{15}, the resulting model thus shows an under-representation of female groups and groups in other age ranges. These biases can be further extended to race, ethnicity and disability \cite{18,19}. As a result, LLMs might generate biased content that instills negative stereotypes and sentiments toward certain demographics \cite{18}, leading to exacerbated social inequality and discrimination.

\paragraph{\textbf{Information inaccuracy.}} Information accuracy is not guaranteed by LLMs, which can end up misleading users who are inexperienced or who otherwise overly trust LLM outputs. LLMs can in fact be negatively impacted by inaccurate information that is inherent in the large corpora used for training them. For example, Reddit as a source of GPT-2 \cite{16} has a risk of containing information that is not verified, leading to potential information inaccuracy and credibility issues. As a result, LLM-generated text might be inaccurate, even if the information is factual \cite{2,15,20}. With LLM-generated content being perceived as a credible source by many users \cite{21}, it might mislead non-professional users with its outputs of varying levels of quality. Meanwhile, if the problem persists, it will lead to a degradation in LLM-generated text accuracy or cause complications in future training of LLMs.

With the massive creation of LLM-generated content on the web, there is a risk that future training of LLMs could include LLM-generated content without necessarily knowing, which can then amplify accuracy issues on the LLMs. Research has suggested that LLM-generated content should be excluded from training to avoid this problem \cite{16} indicating that unnecessary overhead is caused due to the inaccuracy of LLMs. To conclude, considering the increasing usage and potential problems of automatic text generation, it is important to identify LLM-generated text to safeguard the quality of output and to mitigate the aforementioned problems with human oversight.

\subsection{State-of-the-art LLM-generated Text Detection}
\label{ssec:sota-llm-detection}

\paragraph{\textbf{Approaches to LLM-generated text detection.}}
Currently, there are 2 major streams of LLM-generated text detectors: i) zero-shot classifiers \cite{1,2}, and ii) watermark detectors \cite{3}. Zero-shot classifiers aim to identify patterns and statistical characteristics of input text, comparing them to those of LLM-generated text \cite{14,22}. For example, DetectGPT calculates the average log probability ratio of the input text over its perturbations and classifies text as LLM-generated if the ratio exceeds a threshold \cite{2}. Zero-shot classifiers do not require further training, thus facilitating usage from non- technical users. Watermark detectors rely on the addition of a watermark, which is not visible to humans, on LLM-generated text. \cite{3} proposed a watermarking scheme with tokens in ``green list'' and ``red list.'' While adding the watermark on LLM-generated text, the use of ``green list'' tokens are prioritized in the sentence generation process, resulting in a text dominated by these tokens. The detector thus classifies a text to be LLM-generated if the number of ``green list'' tokens is high. Strong and weak watermarking can be implemented by adjusting the parameters \cite{3}. These 2 types of detectors both exhibit remarkable performance in LLM-generated text detection.

For zero-shot classifiers, 95\% accuracy is observed from RoBERTa fine-tuned classifiers \cite{1} and 92\% from GROVER \cite{23}. Apart from accuracy, AUROC is also considered to account for the true positive rate (TPR) and false positive rate (FPR), as false positive is highly discouraged for these classifiers to misclassify human-written text as LLM-generated \cite{3}. DetectGPT shows state-of-the-art performance with an average AUROC of 95.3\% on various datasets. It should also be noted that detection performance varied with text length and decoding strategies \cite{1,2}. Meanwhile, watermark detectors also show outstanding performance, with strong watermarking detection achieving 100\% AUROC and soft watermarking achieving 98.9\% \cite{3}. Watermark detectors are also regarded as more effective detectors than zero-shot classifiers \cite{10}. Other than the above 2 mainstream LLM-generated text detectors, \cite{10} proposed an information retrieval-based detector, which classifies text by comparing the input text with stored LLM outputs.

\paragraph{\textbf{Paraphrase attacks in LLM-generated text detection.}}
While existing LLM-generated text detectors show remarkable performance, they can be vulnerable to paraphrasing attacks, as one can alter generated texts to circumvent detection. Since these detectors perform classification based on the existence of token patterns or watermarks, paraphrasing on LLM-generated text, either by human or AI paraphrasers, can potentially evade the detectors \cite{3,14,22}. Past research has conducted experiments on different types of paraphrasing attacks. \cite{3} performed the attack by replacing words with tokens generated from T5 model and noticed a significant watermark degradation with AUROC decreasing from 99.8\% to 69.6\%. \cite{10} showed that after using DIPPER AI-paraphraser, DetectGPT’s detection rate significantly reduced from 70.3\% to 4.6\% and watermark detection accuracy decreased from 100\% to 57.2\%. The aforementioned research performs only a single round of paraphrasing, and it is sufficient to substantially reduce the detectors' performance. Noticing this, \cite{14} performed recursive paraphrasing attacks on various LLM-generated text detectors, where LLM-generated text is paraphrased with AI paraphrasers in multiple iterations. They concluded that recursive attacks could further evade these detectors. For non-watermarked text, DetectGPT’s AUROC score decreases from 96.5\% to 59.8\% with text paraphrased with T5 model. The same outcome also resulted in watermarked text. With 2 rounds of paraphrasing with DIPPER and Llama-2-7B-Chat \cite{14}, TPR@1\%FPR drops significantly from 99.8\% to 44.8\% and 38.9\% respectively. The score further decreased to 15.7\% and 27.2\% respectively after 5 rounds of paraphrasing, reflecting the detectors' inability to identify LLM-generated paraphrases. Other than the degradation in TPR@1\%FPR, AUROC scores also decreased from 99.9\% to 76.3\% and 79.5\% respectively. As such, from existing research it can be expected that paraphrasing attacks, which change the statistical properties and replace watermarked tokens, could effectively evade both zero-shot classifiers and watermark detectors while preserving semantic information from the original LLM-generated text \cite{10}.

\subsection{Research Gap}
\label{ssec:research-gap}

In existing research, classifications are conducted with datasets consisting of human-written text and paraphrases of LLM-generated text. While LLM-generated text detectors make classification based on the existence of LLM-generated features in the input \cite{14,22}, a potential reason for paraphrasing attack successfully evading these detectors might be the fundamental difference in the language format between LLM-generated text and its paraphrases. \cite{5} stated that paraphrases, which contain similar semantic information, could exhibit different lexical, syntactic and word order from the original text. As such, the LLM-generated paraphrases might not contain the characteristics that these detectors are looking for, leading to misclassification.

Hence, to addressing the existing gap, our aim in this research is to include human-paraphrased text in the dataset for classification, which we experiment with state-of-the-art LLM-generated text detectors. LLM-paraphrased text might potentially distinguish itself from human-paraphrased text, in terms of the degree of semantic understanding, writing style and word choices. As a result, an investigation into the effect of including human paraphrases on LLM-generated text detection can help identify potential reasons for misclassification by existing detectors and to further aid research in developing LLM-generated text detectors.


\section{Dataset}
\label{sec:dataset}

\subsection{Creating the HLPC Dataset}
\label{ssec:creating-dataset}

Despite the existence of numerous datasets containing human- and LLM-generated data, we found no suitable dataset that incorporates also their paraphrases which would enable our study, and hence we leverage and extend existing datasets for creating the HLPC dataset.\footnote{The HLPC dataset can be found at \url{https://github.com/kristylht/Human-LLM-Paraphrase-Collection-HLPC}}

\paragraph{\textbf{Overview}} We put together a dataset consisting of two types of data, each consisting of an original document (DOC) and its paraphrase (PP):

\begin{enumerate}
 \item Human-generated data: collected from 4 existing datasets, which include MRPC, XSum, QQP and MultiPIT, and provide original human-written documents (H-DOC) and their paraphrases (H-PP).
 \item LLM-generated data: original LLM documents (LLM-DOC) are generated by using parts of the H-DOC documents above as prompts sent to the LLM, and their paraphrases (LLM-PP) are generated through a paraphrases that runs five paraphrasing iterations on the LLM-DOC documents.
\end{enumerate}

In what follows we further elaborate to describe the data creation process in detail.

\paragraph{\textbf{Human-written and paraphrased documents (H-DOC and H-PP).}} To retrieve human-generated texts alongside their paraphrases, we make use of four different datasets, namely Microsoft Research Paraphrase Corpus (MRPC) \cite{24}, Extreme Summarization Dataset (XSum) \cite{25}, Quora Question Pairs Dataset (QQP) \cite{26} and Multi-Topic Paraphrases in Twitter (MultiPIT-expert) \cite{27}. MRPC consists of sentence pairs from newswire articles, while XSum dataset contains pairs of BBC articles and their summary. QQP dataset includes question pairs from Quora, and MultiPIT includes pairs of tweets from Twitter. Therefore, the combination of these data sources encompasses both clean and well-structured text data from authoritative news companies and noisy and unstructured text data from online forums. Since these datasets were published before the surge of generative AI, it is assumed that all texts are human-generated. As these datasets were originally used for paraphrase detection, pairs of sentences in these datasets are labelled as paraphrases (label = 1) or non-paraphrases of each other (label = 0), among which we are interested in those with a positive label for our purposes.

\paragraph{\textit{Document filtering.}} From the H-DOC and H-PP samples above, we remove cases of non-paraphrases, sampling only the pairs that are paraphrases to satisfy our objective of having human paraphrases for every human-generated text. Of the remaining pairs, we remove samples with fewer than 10 tokens (fewer than 30 tokens in the case of XSum due to its greater length), to enable prompt extraction for subsequent LLM text generation as described below, as well as those with more than 512 tokens due to restrictions of GPT-2 for text generation. Finally, we randomly sample 150 documents from each set, H-DOC and H-PP, with 300 documents remaining across both types.

\begin{figure}
	\centering
		\includegraphics[scale=.32]{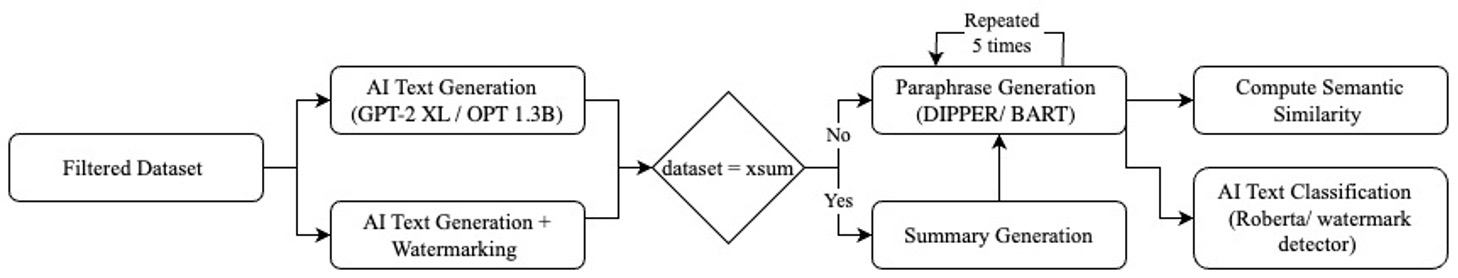}
	\caption{Flowchart of the LLM text generation, paraphrasing and classification process.}
	\label{fig:llm-generation-flowchart}
\end{figure}

We next proceed to generating the LLM texts and their paraphrases. Figure \ref{fig:llm-generation-flowchart} illustrates the ``LLM-generated documents (LLM-DOC)'' and ``LLM-generated paraphrases (LLM-PP)'' generation process.

\paragraph{\textbf{LLM text generation (LLM-DOC).}}
To generate LLM-based texts that resemble the human-generated texts above, we first obtain prompts from the human-generated texts above, which are then fed to the LLM to generate new texts. The LLM-DOC generation process starts by taking the first 5 tokens from H-DOC (first 30 tokens for documents sourced from XSum) with a tokenizer, which are used as prompts for the language models in the generation process. The models then generate text based on the given prompt up to a length of the maximum length of the H-DOC. We generate two types of LLM-DOC, watermarked and non-watermarked, both using two transformer-based language models, GPT2-XL \cite{1} and OPT-1.3B \cite{9}, along with their respective tokenizers. These models are pretrained with a wide range of internet text and therefore are suitable for text generation in this project. Default settings are used for parameter initialization for both models. Non-watermarked LLM-generated texts are output purely generated from the above two models. For watermarked LLM-DOC generation, watermarks are added by initializing a watermark in the logit processor of the models in addition to the LLM-DOC process \cite{3}.

\paragraph{\textbf{LLM paraphrase generation (LLM-PP).}}
We then use the LLM-generated documents above to generate their paraphrases. For the outputs from the XSum dataset, summaries are first generated using a fine-tuned T5 model specializing in news summarisation \cite{28}. The output summary is limited to a length of maximum length of H-PP in the original XSum dataset. The summaries, along with the LLM-DOC generated documents from the other 3 datasets, are taken to generate LLM-PP. We use two paraphrasers, namely DIPPER and BART-paraphrase models. DIPPER is a T5-XXL paraphrase generation model with 11 billion parameters, fine-tuned on 6.3 million data points. Inherently, DIPPER is capable of capturing long-term dependencies and controlling output diversity \cite{10}. However, due to computational resource limitations, we use a non-context version, resulting in reduced performance on long-term dependency capturing. To generate paraphrases with models that capture long-range dependencies, the second model, BART is used. The model is built upon a seq2seq architecture, with a bidirectional encoder and a unidirectional decoder \cite{29}. The bidirectional encoder allows the model to understand sentence embeddings in a longer range, thus providing more semantic and contextual information for later paraphrase generation. Particularly, the BART-paraphrase model used in this project is a fine-tuned BART model pretrained with 3 paraphrase datasets, providing better performance in paraphrase generation. The LLM-DOCs from each dataset are passed to these paraphrasers for paraphrase (LLM-PP) generation. In order to investigate the effect of the recursive paraphrasing attack mentioned in \cite{14}, 5 rounds of paraphrase generation are conducted, with the 1st of paraphrases generated from the LLM-DOC, and the subsequent rounds of paraphrases generated iteratively using the outputs of each round.

\paragraph{\textbf{Final dataset.}} The final HLPC dataset is composed of 600 documents, with a balanced distribution of 150 documents per type, i.e. H-DOC, H-PP, LLM-DOC and LLM-PP. These documents are grouped into two categories, human-generated (H-DOC and H-PP) and LLM-generated (LLM-DOC and LLM-PP). We use these two categories to perform binary classification in the LLM- vs human-generated text detection task, where we mix both original and paraphrased documents to evaluate their impact (particularly that of human-written paraphrases H-PP) on the detection models.

\subsection{Evaluating the quality of LLM-generated paraphrases}
\label{ssec:quality-paraphrases}

Since paraphrases should inherently not deviate much from the original text, it is essential to evaluate the quality of the paraphrases in terms of semantic and contextual preservation. Therefore, we evaluate the semantic similarities between the original texts and their paraphrases to assess the quality of the paraphrasing. We use an automated, fine-tuned sentence transformer model \cite{30} to generate sentence embeddings for each text-paraphrase pair, and we calculate the cosine similarity between the embeddings to account for semantic similarity, scoring from -1 (least similar) to 1 (most similar). Semantic similarities are calculated between H-DOC and H-PP and between LLM-DOC and each round of LLM-PP.

\begin{figure}
	\centering
	\begin{subfigure}{0.4\linewidth}
		\includegraphics[width=\linewidth]{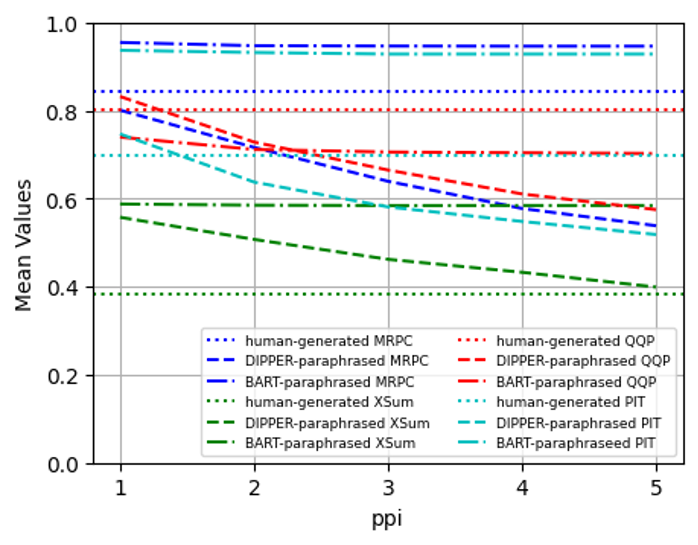}
		\label{fig.2a}
	\end{subfigure}
	\begin{subfigure}{0.4\linewidth}
		\includegraphics[width=\linewidth]{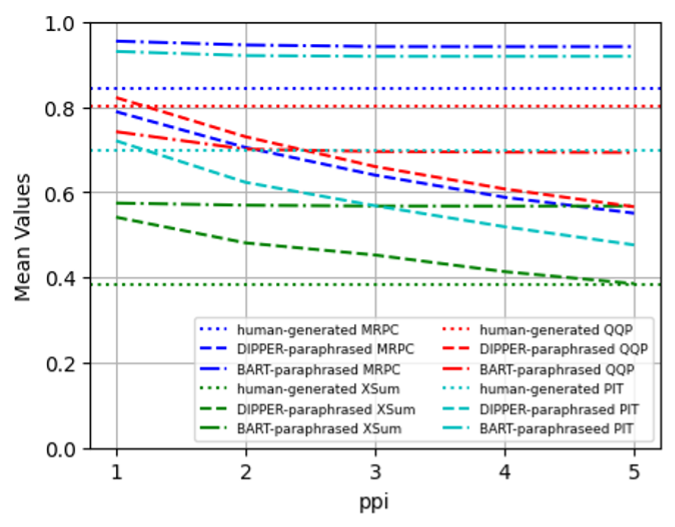}
		\label{fig.2b}
	\end{subfigure}
	\caption{Mean similarity scores of watermarked (left) and non-watermarked (right) LLM-generated texts and their paraphrases across different datasets.}
	\label{fig:paraphrase-similarities}
\end{figure}

Looking at human-generated data, MRPC, QQP and MultiPIT score over 0.7 for mean semantic similarity scores, indicating a good level of semantic preservation, while XSum scores only 0.383 since paraphrases in XSum are summaries of the documents. When it comes to LLM-generated data, first, the results from paraphrases generated from DIPPER and BART are compared. Figure \ref{fig:paraphrase-similarities} shows the similarity scores of watermarked and non-watermarked LLM-DOC and LLM-PP. From both graphs, BART outperforms DIPPER in all datasets, particularly in MRPC and MultiPIT where BART scores over 0.9 across paraphrasing rounds which is even higher than the score from human-generated data. DIPPER’s performances are generally worse than human paraphrasing, except in XSum. Meanwhile, paraphrases from both paraphrasers exhibit degradation in similarity scores across paraphrasing rounds. With recursive paraphrasing, similarity scores from BART decrease slightly from an average mean score of 0.80 in the 1st round of paraphrasing to 0.785 in the 5th round, while similarity scores of paraphrases generated by DIPPER decrease significantly from an average mean score of 0.727 in the 1st round of paraphrasing to 0.501 in the 5th round. BART’s performance under recursive paraphrasing is within expectation since paraphrases generated across rounds are similar. However, the significant degradation in semantic similarity with DIPPER-generated paraphrases indicates that much semantic information is lost across paraphrasing rounds, resulting in poor-quality paraphrases that deviate from the original LLM-DOC. Second, the results from paraphrases generated from watermarked and non-watermarked LLM-DOC are compared. From Figure \ref{fig:paraphrase-similarities}, both semantic scores of paraphrases datasets, with average mean semantic scores of 0.732 and 0.746, and average degradation rates of mean semantic scores of 0.168 and 0.162 respectively. Therefore, it can be concluded that watermarking does not affect the quality of paraphrases.

Lastly, we compare the results from paraphrases generated from GPT2-generated and OPT-generated documents. Paraphrases generated from GPT2-generated documents are of higher quality, with an average mean semantic score of 0.788 compared to 0.670 from OPT-generated documents. Meanwhile, the rate of quality degradation is similar for both models, with average degradation rates of mean semantic scores of 0.157 for GPT2 and 0.160 for OPT.

\begin{table}[h]
    \caption{Examples of LLM-generated documents.}

    \begin{tabular}{l|p{0.15\textwidth}|p{0.6\textwidth}}
         Original Text & \multicolumn{2}{p{0.75\textwidth}}{The dollar rose 0.6 percent to 109.54 yen < JPY = > and climbed more than 1 percent to 1.3315 Swiss francs < CHF = >.} \\ 
         \hline
         Non-watermarked & GPT Output &The dollar rose 0.3 percent to \$1.3240 per euro, while the euro was little changed at \$1.3240. The dollar index \\
         \cline{2-3}
         & OPT Output &The dollar rose 0.2 percent against the yen to 111.85 yen, after hitting a two-week high of 111.90 yen on Friday. \\ 
         \hline
         Watermarked & GPT Output & The dollar rose 0.2 percent to \$1.1234 from \$1.1218, after touching a high of \$1.1218 on Friday. \\ 
         \cline{2-3}
         & OPT Output &The dollar rose 0.1 percent against the yen to 110.65 yen, after hitting a three-month high of 110.78 yen on Tuesday.
    \end{tabular}
    \label{table.1}
\end{table}

\subsection{Descriptive Statistics of HLPC}
\label{ssec:descriptive-stats}

\begin{table}[b]
\caption{Examples of LLM-generated Paraphrases.}
    \begin{tabular}{p{1.5cm}|c|p{13cm}}
        Paraphraser & ppi & Text \\ 
        \hline 
        \multirow{6}{2cm}{DIPPER} & i = 0 & The dollar rose 0. 2 percent to \$1.1234 from \$1.1218, after touching a high of \$1.1218 on Friday. \\ 
        & i = 1 & The dollar rose 0.10 percent to \$1.1234 from \$1.1218, after a high of \$1.1218 on Friday \\ 
        & i = 2 & The dollar rose 0.10 percent to \$1.1234 from \$1.1218. \\ 
        & i = 3 & The dollar rose by a penny to \$1.1234 from \$1.1218 \\ 
        & i = 4 & The dollar jumped a penny to \$1.1234 from \$1.1218. \\ 
        & i = 5 & A little more, the dollar was up a penny to \$1.1234. \\ 
        \hline 
        \multirow{6}{2cm}{BART} & i = 0 & The dollar rose 0. 2 percent to \$1.1234 from \$1.1218, after touching a high of \$1.1218 on Friday. \\ 
        & i = 1 & The dollar rose 0.2 percent to \$1.1234 on the New York Stock Exchange, after touching a high of \$.1218 on Friday. \\ 
        & i = 2 & The dollar rose 0.2 percent to \$1.1234 on the New York Stock Exchange, after touching a high of \$.1218 on Friday. \\ 
        & i = 3 & The dollar rose 0.2 percent to \$1.1234 on the New York Stock Exchange, after touching a high of \$.1218 on Friday. \\ 
        & i = 4 & The dollar rose 0.2 percent to \$1.1234 on the New York Stock Exchange, after touching a high of \$.1218 on Friday. \\ 
        & i = 5 & The dollar rose 0.2 percent to \$1.1234 on the New York Stock Exchange, after touching a high of \$.1218 on Friday.
    \end{tabular}
    
    \label{table.2}
\end{table}

Of the four data sources that we use to build HLPC, documents and paraphrases are shorter with MRPC, QQP and MultiPIT, with the highest mean length of 23.19 from MRPC documents, and the lowest mean length of 12.41 from MultiPIT paraphrases. XSum has longer passages as documents, having a mean length of 269.23, and its paraphrases of 22.09.

Looking at the LLM-generated outputs in our dataset, generally GPT2 produces outputs with longer text length, having an average mean length of 123.54 for non-watermarked outputs and 124.53 for watermarked outputs, compared to 117.95 for OPT non-watermarked outputs and 118.55 for OPT watermarked outputs. Meanwhile, watermarking does not affect text length significantly, with watermarked outputs having a slightly higher average mean length (121.54) than non-watermarked outputs (120.75). This is because watermarking does not directly decide token choices during text generation, it simply promotes the probability of certain tokens. An example is shown in Table \ref{table.1} that watermarking influences the choice of tokens but poses minimal effects on text length.

We next look at the LLM paraphrases (LLM-PP). First, we observe that the text length of paraphrases decreases as the number of paraphrase rounds increases. This is more obvious from paraphrases generated by DIPPER, with a 31\% decrease in average mean length from 23.82 in the 1st round of paraphrases to 16.38 in the 5th round. Paraphrases generated from BART are similar in text length across rounds of paraphrases. There is no significant difference between the text length of paraphrases generated from watermarked and non-watermarked LLM-DOC. Second, in terms of content diversity, DIPPER generates paraphrases with different choices of wordings compared to the original text while preserving the semantic information, and BART generates paraphrases that are similar or even identical to the original text. An example of recursive DIPPER- and BART-generated paraphrases of watermarked GPT output from MRPC is presented in Table \ref{table.2}. For more such examples, please see Appendix \ref{app:llm-examples}.

\section{LLM-generated Text Detection Experiments}
\label{sec:llm-generated-detection}

We next describe the LLM-generated text detection models we use for our experiments, as well as the evaluation metrics.

\paragraph{\textbf{LLM-generated text detection models.}} For our experiments, we choose to use two state-of-the-art models as LLM-generated text detectors, namely OpenAI RoBERTa Detector \cite{1} and watermark detector \cite{3}. The OpenAI Detector is a fine-tuned RoBERTa model trained with outputs from GPT2 model and thus is able to detect GPT2 and various LLM output text \cite{1}. The watermark detector classifies text by computing the number of green tokens and the probability of its existence in the given input \cite{3}. The given input is classified as LLM-generated if the probability exceeds the set threshold.

Using the HLPC dataset, we test models on a balanced set of 600 documents, 300 human-generated and 300 LLM-generated. For testing involving LLM-PP, the experiment is repeated 5 times, using each of the 5 rounds of AP. The parameters of the watermark detector are set according to the parameters used in watermarked AI document generation for effective classification. OpenAI Detector is used on non-watermarked LLM-DOC and their paraphrases, while the watermark detector on watermarked LLM-DOC and their paraphrases.

\paragraph{\textbf{Experiment settings.}} The experiment is repeated with the outputs from different combinations of the 4 datasets (MRPC, XSum, QQP and MultiPIT), 2 generative language models (GPT and OPT) and 2 paraphrasers (DIPPER and BART). The final classification is conducted with the full set of human-generated data and LLM-generated data, which means that 150 samples are taken from each of the documents and paraphrases, resulting in a total sample size of 600 for classification.

\paragraph{\textbf{Evaluation metrics.}} To evaluate the performance of the LLM text detectors, we compute the accuracy, TPR@1\%FPR and AUROC for each round of classification. These metrics are chosen as they are widely used in the literature, and they provide a comprehensive analysis of the classification performance. Specifically, AUROC provides an overview of the tradeoff between the true positive rate (correctly classifying LLM-generated data) and false positive rate (misclassifying human-generated data as LLM-generated), and TPR@1\%FPR shows the performance of the classifiers focused on improving the True Positive Rate (TPR) while also trying to keep the False Positive Rate (FPR) low.

The equations of AUROC, TPR and FPR are as follows:

\begin{equation}
    \label{eq:trp}
    \text{TPR} = \frac{\text{True Positives (TP)}}{\text{True Positives (TP)} + \text{False Negatives (FN)}} \quad \text{FPR} = \frac{\text{False Positives (FP)}}{\text{False Positives (FP)} + \text{True Negatives (TN)}}
\end{equation}

\begin{equation}
    \label{eq:auroc}
    \text{AUROC} = \int_0^1 \text{TPR}(t) \, d\text{FPR}(t)
\end{equation}

TPR@1\%FPR is the TPR while FPR is set to a 1\% threshold. To investigate the effects of H-PP in LLM-generated text classification, comparisons of classifiers’ performance are made based on the difference in the above statistics from various text pairs. Results between LLM-DOC and LLM-PP when paired with H-DOC and H-PP are compared. 

\section{Results \& Analysis}
\label{sec:results}

\subsection{Classification with human-generated data and LLM-generated documents}
\label{ssec:classification-human-llm}

We perform two classification experiments to observe the effects of including H-PP on LLM-generated text detection. The first classification is conducted with H-DOC and LLM-DOC, while the second classification with H-PP and LLM-DOC.

\begin{figure}[htb]
	\centering
        \begin{subfigure}{0.42\linewidth}
            \includegraphics[width=\linewidth]{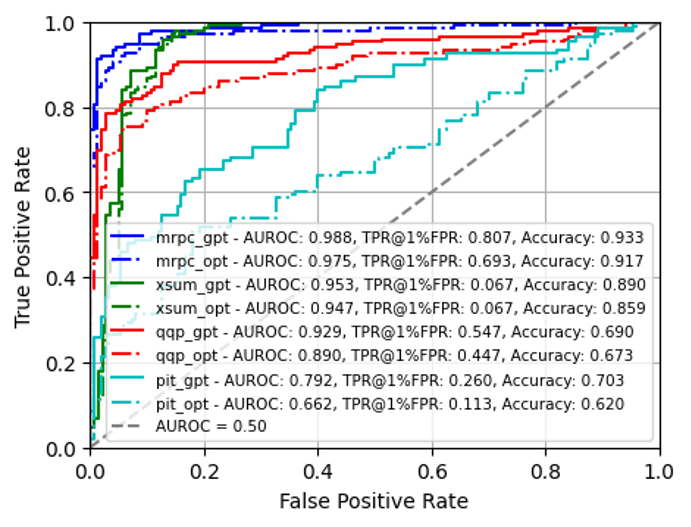}
            \subcaption{}
            \label{fig.3a}
        \end{subfigure}
        \begin{subfigure}{0.42\linewidth}
            \includegraphics[width=\linewidth]{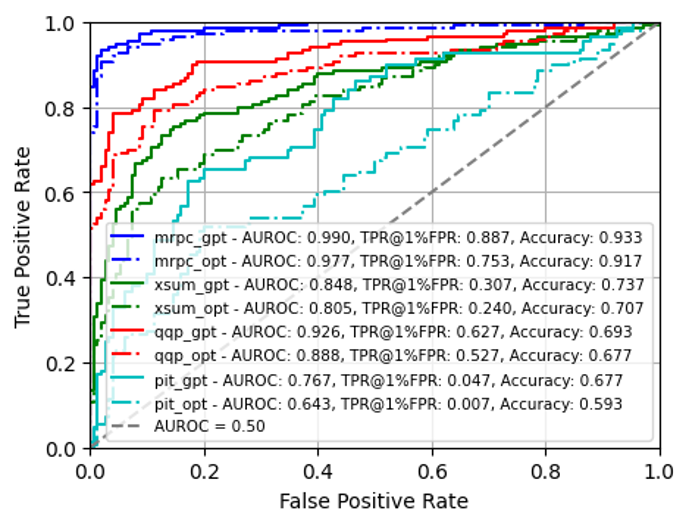}
            \subcaption{}
            \label{fig.3b}
        \end{subfigure}
        \begin{subfigure}{0.42\linewidth}
            \includegraphics[width=\linewidth]{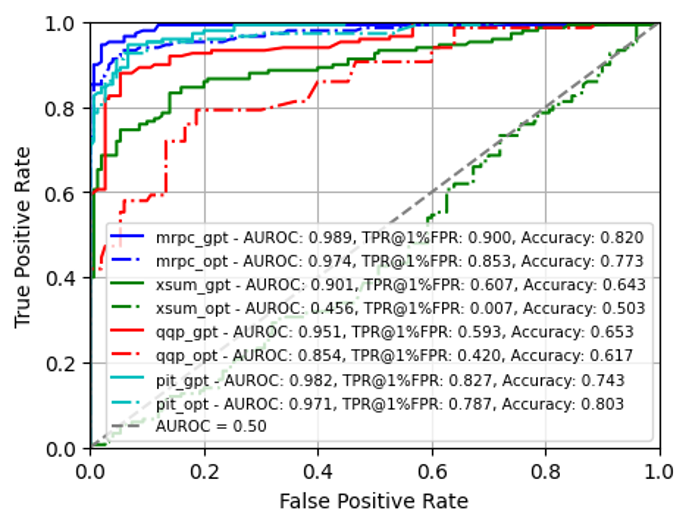}
            \subcaption{}
            \label{fig.3c}
        \end{subfigure}
        \begin{subfigure}{0.42\linewidth}
            \includegraphics[width=\linewidth]{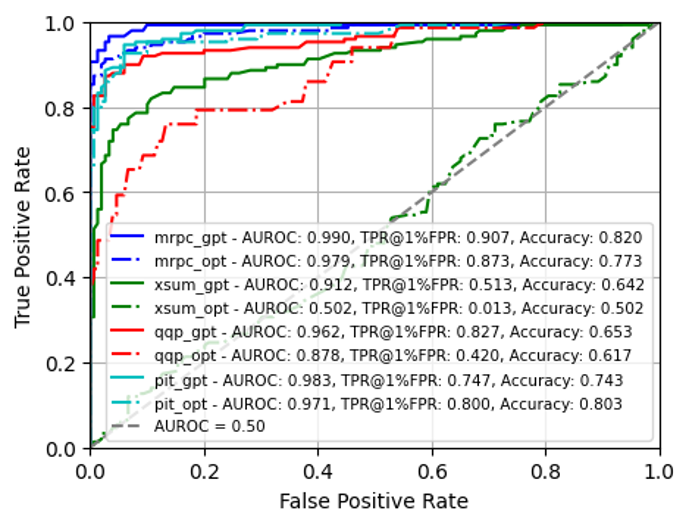}
            \subcaption{}
            \label{fig.3d}
        \end{subfigure}

	\caption{ROC Curve Comparison: (a) Human-generated Documents vs Non-watermarked LLM-generated Documents, (b) Human-generated Paraphrases vs Non-watermarked LLM-generated Documents, (c) Human-generated Documents vs Watermarked LLM-generated Documents, (d) Human-generated Paraphrases vs Watermark LLM-generated Documents}
	\label{fig.3}
\end{figure}

Figure \ref{fig.3a} shows the ROC curve of the results with non-watermarked and watermarked LLM-DOC respectively. First, the results of classifications in terms of non-watermarked and watermarked LLM-DOC are evaluated. For both non-watermarked and watermarked LLM-DOC, the results are satisfactory either with or without H-PP, with over 75\% of classifications scoring an AUROC > 0.85. Among them, the results from watermarked LLM-DOC are better than the results from non-watermarked LLM-DOC, with 87.5\% of classification scoring an AUROC >0.85, compared to 62.5\% from non-watermarked LLM-DOC. This shows that watermarking is a more effective strategy in LLM-generated text detection.

Figures \ref{fig.3a} and \ref{fig.3b} show that the inclusion of H-PP generally decreases AUROC and accuracy and increases TPR@1\%FPR, compared to the results with H-DOC. Among all the data sources, results from Xsum show the highest decrease of 0.142 AUROC and 0.153 accuracy and the highest increase of 0.24 TPR@1\%FPR. This implies that H-PP might contain similar semantic and contextual information to LLM-DOC, making it more challenging for the model to distinguish between the classes. However, the increase in TPR@1\%FPR indicates the identification of LLM-DOC is promoted while ensuring a low percentage of misclassification of human-generated data as LLM-generated (False Positive) with H-PP.

In addition, Figures \ref{fig.3c} and \ref{fig.3d} show a different phenomenon while including H-PP with watermarked LLM-DOC, with the inclusion of H-PP generally increasing AUROC, TPR@1\%FPR, and posing no effects on accuracy. Among all the data sources, Xsum shows the highest increase of 0.046 AUROC, and QQP shows the highest increase of 0.234 TPR@1\%FPR. This implies that model performances are promoted while H-PP is included with watermarked LLM-DOC. Overall, the results show the effects of including H-PP are highly dependent on the types of LLM- DOC used in classification. While it increases TPR@1\%FPR in all scenarios, it also decreases AUROC and accuracy when non-watermarked LLM-DOC are used. 

\subsection{Classification under Recursive Paraphrasing}
\label{ssec:classification-recursive-paraphrasing}

We next evaluate the effects of including H-PP with recursive paraphrases. In the interest of brevity and focus, we analyze results for MRPC GPT-generated text here, but provide full details for all datasets in Appendix \ref{appendix a.3}, whose results are consistent with MRPC. The results (H-PP\_pp0, H-PP\_pp2 \& H-PP\_pp5) are compared to the classification result without the use of H-PP (pp0, pp2 \& pp5). First, the general model performance is discussed in terms of the differences between DIPPER and BART-generated paraphrases.

\begin{figure}
	\centering
        \begin{subfigure}{0.42\linewidth}
            \includegraphics[width=\linewidth]{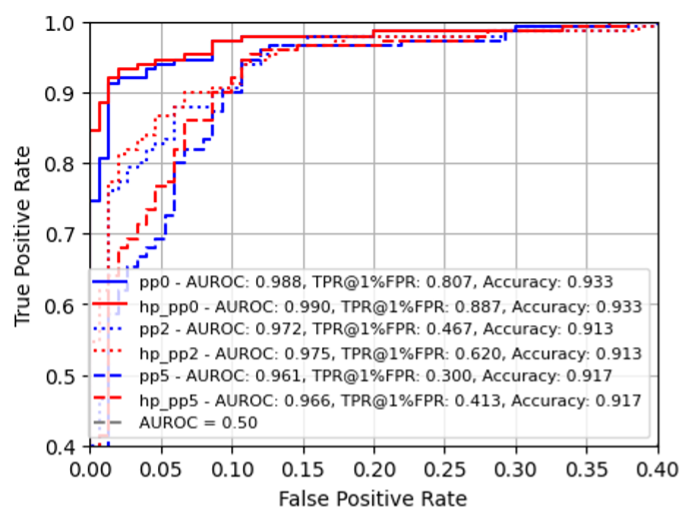}
            \subcaption{}
            \label{fig.4a}
        \end{subfigure}
        \begin{subfigure}{0.42\linewidth}
            \includegraphics[width=\linewidth]{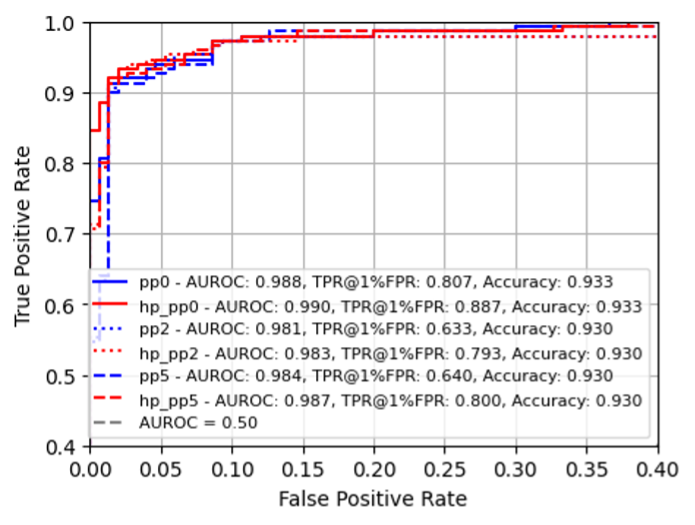}
            \subcaption{}
            \label{fig.4b}
        \end{subfigure}
        \begin{subfigure}{0.42\linewidth}
            \includegraphics[width=\linewidth]{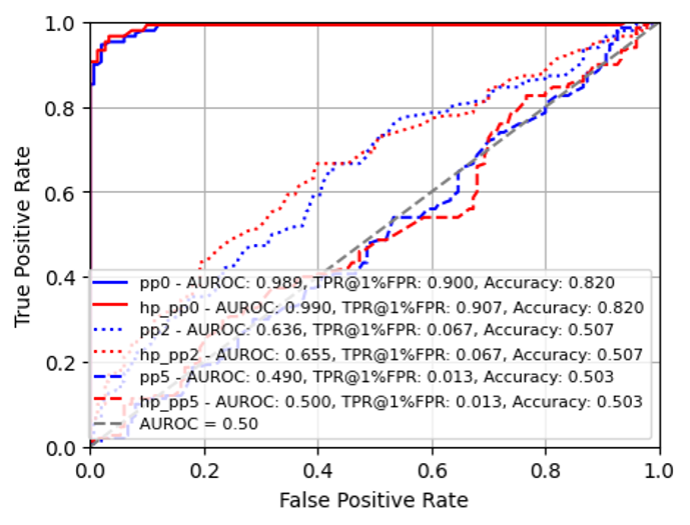}
            \subcaption{}
            \label{fig.4c}
        \end{subfigure}
        \begin{subfigure}{0.42\linewidth}
            \includegraphics[width=\linewidth]{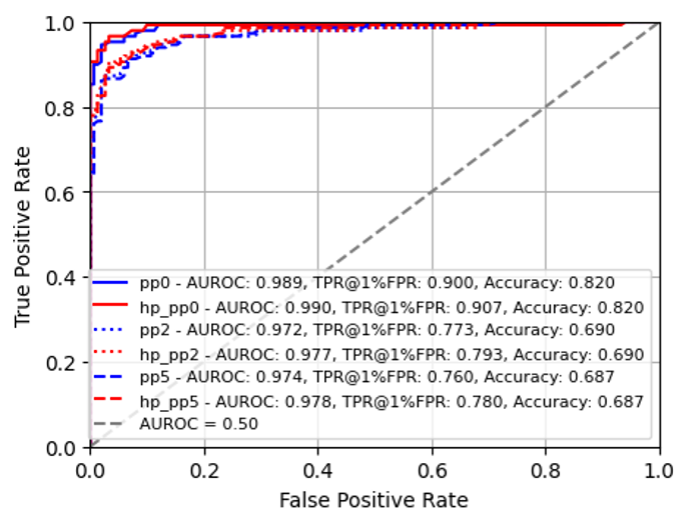}
            \subcaption{}
            \label{fig.4d}
        \end{subfigure}

	\caption{ROC Curve Comparison: (a) Human-generated Documents/Paraphrases vs DIPPER-generated Paraphrases from Non-watermarked MRPC GPT2-generated Text, (b) Human-generated Documents/Paraphrases vs BART-generated Paraphrases from Non-watermarked MRPC GPT2-generated Text, (c) Human-generated Documents/Paraphrases vs DIPPER-generated Paraphrases from Watermarked MRPC GPT2-generated Text, (d) Human-generated Documents/Paraphrases vs BART-generated Paraphrases from Watermarked MRPC GPT2-generated Text}
	\label{fig.4}
\end{figure}

Figure \ref{fig.4} shows that regardless of the presence of watermarking and H-PP, model performances with DIPPER-generated paraphrases degrade to a larger extent than with BART-generated paraphrases across rounds of paraphrasing. A decrease of 0.26 in average AUROC and 0.67 in average TPR@1\%FPR are observed from DIPPER, while BART’s average AUROC and TPR@1\%FPR decrease merely 0.0075 and 0.045 respectively. Particularly, the classification with DIPPER-generated paraphrases from watermarked LLM-DOC (Figure \ref{fig.4c}) shows the highest degradation in model performance, resulting in the lowest AUROC of 0.49 which is worse than a random classifier. This can be attributed to the lower semantic similarity between the LLM-DOC and DIPPER-generated LLM-PP mentioned in Section \ref{ssec:quality-paraphrases}. The low semantic similarity might indicate that DIPPER-generated LLM-PP becomes more similar to H-PP coincidentally while deviating from the original LLM-DOC, resulting in a degradation in model performance.

In addition, classification with BART-generated LLM-PP (Figure \ref{fig.4d}) shows excellent results with AUROC \~0.98 even after 5 rounds of paraphrasing. Second, the effects of including H-PP in classification are evaluated. In both classifications with paraphrases from watermarked and non-watermarked LLM-DOC, the performances of the models with H-PP are better than those without H-PP, while the effect of the promotion is more significant with paraphrases generated from non-watermarked LLM-DOC. For classification with paraphrases generated from non-watermarked LLM-DOC and H-PP (Figures \ref{fig.4a} and \ref{fig.4b}), TPR@1\%FPR increases from a minimum of 0.02 to a maximum of 0.153 when compared to classification with H-DOC. The positive effect is less significant on AUROC with an average increase of 0.0028. Although minimal effects are cast on AUROC and accuracy with the use of H-PP, the significant increase in TPR@1\%FPR shows that H-PP helps in the detection of LLM-generated data at a low false positive rate, ensuring that less human data are classed as LLM-generated.

For classification with paraphrases generated from watermarked LLM-DOC and H-PP (Figures \ref{fig.4c} and \ref{fig.4d}), minimal increases in AUROC and TPR@1\%FPR are observed, with the highest increase in AUROC of 0.019 and TPR@1\%FPR of 0.02. Overall, it can be concluded that including H-PP in the classification under recursive paraphrasing helps promoting AUROC and TPR@1\%FPR under recursive paraphrasing. The comparison of classification results between the inclusion and exclusion of H-PP with other datasets and generative models are also presented in Appendix \ref{appendix a.3}. Generally, similar results are observed, except for classification with Xsum where the inclusion of H-PP reduces AUROC.

\subsection{Classification with Full Set of Human-generated Data and LLM-generated Data}
\label{ssec:classification-full-set/}

\begin{table}[h]
    \caption{Average results of classification with full human-written and LLM-generated data; bracketed numbers indicate percentage difference.}
    \raggedright
    \begin{tabular}{p{0.1\textwidth}|p{0.09\textwidth}|l|l|l|l}
         LLM-generated Documents & Paraphraser & Data & AUROC & TPR\@1\%FPR & Accuracy \\ 
    \hline 
        \multirow{12}{2cm}{Non-watermarked} & \multirow{6}{2cm}{DIPPER} & pp1 & \textbf{0.908} & 0.285 & \textbf{0.814} \\ 
        && hp\_pp1 & 0.883 & 0.358 & 0.769 \\
        && f\_pp1 & 0.884 (-2.71\%) & \textbf{0.379 (32.98\%, 5.86\%)} & 0.778 (-4.62\%) \\
        && \cellcolor{gray!40}pp5 & \cellcolor{gray!40}\textbf{0.878} & \cellcolor{gray!40}0.138 & \cellcolor{gray!40}\textbf{0.811} \\
        && \cellcolor{gray!40}hp\_pp5 & \cellcolor{gray!40}0.843 & \cellcolor{gray!40}0.169 & \cellcolor{gray!40}0.766 \\
        && \cellcolor{gray!40}f\_pp5 & \cellcolor{gray!40}0.867 (-1.27\%) & \cellcolor{gray!40}\textbf{0.290 (110.14\%, 71.60\%)} & \cellcolor{gray!40}0.775 (-2.32\%) \\ 
    \cline{2-6}
        & \multirow{6}{2cm}{BART} & pp1 & 0.873 & 0.209 & 0.793 \\ 
        && hp\_pp1 & 0.839 & 0.284 & 0.750 \\
        && f\_pp1 & \textbf{0.874 (0.001\%, 4.17\%)} & \textbf{0.313 (49.76\%, 10.21\%)} & \textbf{0.771 (-2.85\%)} \\
        && \cellcolor{gray!40}pp5 & \cellcolor{gray!40}\textbf{0.876} & \cellcolor{gray!40}0.201 & \cellcolor{gray!40}\textbf{0.795} \\
        && \cellcolor{gray!40}hp\_pp5 & \cellcolor{gray!40}0.842 & \cellcolor{gray!40}0.276 & \cellcolor{gray!40}0.751 \\
        && \cellcolor{gray!40}f\_pp5 & \cellcolor{gray!40}0.874 (-0.002\%) & \cellcolor{gray!40}\textbf{0.341 (69.65\%, 23.55\%)} & \cellcolor{gray!40}0.773 (-2.85\%) \\ 
    \hline 
     \multirow{12}{2cm}{Watermarked} & \multirow{6}{2cm}{DIPPER} & pp1 & 0.652 & 0.100 & 0.508 \\ 
        && hp\_pp1 & \textbf{0.678} & 0.117 & 0.508 \\
        && f\_pp1 & 0.636 (-6.60\%) & \textbf{0.209 (109\%, 78.63\%)} & \textbf{0.556 (9.45\%, 9.45\%)} \\
        && \cellcolor{gray!40}pp5 & \cellcolor{gray!40}0.512 & \cellcolor{gray!40}0.019 & \cellcolor{gray!40}0.500 \\
        && \cellcolor{gray!40}hp\_pp5 & \cellcolor{gray!40}0.541 & \cellcolor{gray!40}0.022 & \cellcolor{gray!40}0.500 \\
        && \cellcolor{gray!40}f\_pp5 & \cellcolor{gray!40}\textbf{0.587 (14.65\%, 8.5\%)} & \cellcolor{gray!40}\textbf{0.185 (873.68\%, 740.91\%)} & \cellcolor{gray!40}\textbf{0.554 (10.8\%, 10.8\%)} \\ 
    \cline{2-6}
        & \multirow{6}{2cm}{BART} & pp1 & 0.825 & 0.445 & \textbf{0.611} \\ 
        && hp\_pp1 & \textbf{0.840} & \textbf{0.467} & \textbf{0.611} \\
        && f\_pp1 & 0.706 (-18.98\%) & 0.315 (-48.25\%) & 0.584 (-4.62\%) \\
        && \cellcolor{gray!40}pp5 & \cellcolor{gray!40}0.797 & \cellcolor{gray!40}0.398 & \cellcolor{gray!40}\textbf{0.599} \\
        && \cellcolor{gray!40}hp\_pp5 & \cellcolor{gray!40}\textbf{0.813} & \cellcolor{gray!40}\textbf{0.426} & \cellcolor{gray!40}0.598 \\
        && \cellcolor{gray!40}f\_pp5 & \cellcolor{gray!40}0.700 (-16.14\%) & \cellcolor{gray!40}0.307 (-38.76\%) & \cellcolor{gray!40}0.581 (-3.10\%) 
 \end{tabular}
    \label{table.3}
\end{table}

We now perform classification experiments with the full set of human-written and LLM-generated data, passing both documents and paraphrases to the classifier.

Table \ref{table.3} shows the comparison of the average results from classification with i) H-DOC vs LLM-PP (pp1 \& pp5), ii) H-PP vs LLM-PP (H-PP\_pp1 \& H-PP\_pp5) and iii) full human data vs LLM-generated data (f\_pp1 \& f\_pp5) across all datasets. First, for results from paraphrases generated from watermarked LLM-DOC, model performance shows extreme opposites depending on the paraphrasers used. The best improvement in statistical results is shown using DIPPER-generated paraphrases from watermarked LLM-DOC, along with H-DOC, H-PP and LLM-DOC (f\_pp1 \& f\_pp5). With the 1st round of paraphrases, TPR@1\%FPR increases by 109\% and 78.63\% and accuracy increases by 9.45\% and 9.45\%, compared to only using H-DOC or H-PP respectively. With the 5th round of paraphrases, AUROC increases by 14.65\% and 8.5\%, TPR@1\%FPR increases by 874\% and 740\% and accuracy increases by 10.8\% and 10.8\%, compared to only using H-DOC or H-PP respectively. This shows that using the full set of data as input for classification is significantly effective in improving LLM-generated text detection under recursive paraphrasing, with the condition that the LLM-DOC is watermarked and paraphrases are DIPPER-generated. Meanwhile, the worst performance is shown with BART-generated paraphrases from watermarked LLM-DOC under the same condition of using the full set of data as input. Compared with the best results, results from 1st and 5th rounds of BART-generated paraphrases show a degradation of 18.98\% and 16.14\% in AUROC, 48.25\% and 38.76\% in TPR@1\%FPR and 4.62\% and 3.10\% in accuracy respectively. The significant difference in the statistical results shows that the watermark detector is highly sensitive to the paraphraser used. While the performance improves significantly with DIPPER-generated paraphrases, it also degrades significantly with BART-generated paraphrases.

Second, TPR@1\%FPR significantly increases while using paraphrases generated from non-watermarked LLM-DOC along with H-DOC, H-PP and LLM-DOC, while AUROC and accuracy remain unchanged or decrease slightly. Results show that TPR@1\%FPR increases by a minimum of 5.86\% to a maximum of 110.14\%, with an average of 42.84\% across different inputs. This indicates that having the full set of data as input effectively improves TPR@1\%FPR and ensures minimal misclassification at a low false positive rate. Meanwhile, AUROC and accuracy decrease, compared to the results from H-DOC vs LLM-PP and H-PP vs LLM-PP. However, the decrease is less significant compared to the increase in TPR@1\%FPR. The maximum decrease in AUROC and accuracy is merely 4.17\% and 4.62\% respectively. Therefore,  it can be concluded that having a full set of data as input exhibits a trade-off between accuracy, AUROC and TPR@1\%FPR. Considering the significant improvement in TPR@1\%FPR and the importance of ensuring minimal misclassification, having a full set of data as input is suggested to be a better method than using merely H-DOC under recursive paraphrasing. Detailed ROC curve and statistical results of classification with each dataset and its generations are presented in Appendix \ref{appendix a.4}.

\subsection{Benefits of Using Human-written Paraphrases in LLM or Detectors Training}

In our review of the literature in Section \ref{sec:related-work} we found that existing LLMs are trained on corpora that do not contain H-PP information. Since existing detectors are designed to identify the LLM-generated statistical pattern and watermark from the input text, paraphrasing which reduces or erases the above characteristics could effectively evade the detectors. Our experiments show that including H-PP in the dataset promotes classification performances under different circumstances, and therefore, including H-PP in the training datasets during the training of detectors could effectively improve detectors' classification performances since models could learn about the fundamental differences of semantic and contextual information between human-written and LLM-generated text, even under recursive paraphrasing.

Our results also show that the effectiveness of including H-PP in the dataset is highly dependent on the existence of watermarking and the type of paraphraser used. In our experiments, while H-DOC is included for classification, watermarking and DIPPER-generated paraphrases help improve classification performance, while experiments with non-watermarked and BART-generated paraphrases show the opposite. As such, it is important to understand and predict the potential usage of watermarking and the type of paraphraser while developing the detectors. Detector developers could either get the information from users or employ a multi-step classification model for accurate prediction. A multi-step classification model could first identify the presence of watermark and the type of paraphrasers, then decide whether to include H-PP in the training dataset of the detectors based on the results. If such technology or information is not available, it is recommended to include either H-DOC or H-PP, to avoid significant degradation in classification performance. Meanwhile, we show that including H-PP in the datasets is highly effective under recursive paraphrasing. As such, we recommend that detectors, which are used in circumstances where paraphrasing is prevalent, for example, in academic publications, to be trained with H-PP instead of only H-DOC, so as to increase AUROC and TPR@1\%FPR.

\section{Conclusion}
\label{sec:conclusion}

In this study, our aim was to investigate the effect of human paraphrases (H-PP) on LLM-generated text detection by conducting classifications with various combinations of human and LLM-generated data pairs. To enable this study, we devise a data collection strategy and generate the HLPC dataset by leveraging and extending four existing data sources: MRPC, XSum, QQP and MultiPIT. Unlike previous datasets, our new dataset, Human \& LLM Paraphrase Collection (HLPC), incorporates human-written documents (H-DOC), human-written paraphrases (H-PP), LLM-generated texts (LLM-DOC) and LLM-generated paraphrases (LLM-PP). We generate LLM documents by prompting GPT2-XL and OPT-13B with prompts derived from human-written documents. AI paraphraser, DIPPER and BART are then used to paraphrase the generated outputs.
Using this dataset, we perform classification experiments with state-of-the-art LLM-generated text detectors OpenAI RoBERTa and watermark detector, with the aim of understanding the effects of incorporating human-written paraphrases in LLM-generated text detection. Data pairs used for classifications include i) H-DOC vs LLM-DOC, ii) H-PP vs LLM-DOC, iii) H-DOC vs LLM-PP, iv) H-PP vs LLM-PP and v) H-DOC \& H-PP vs LLM-DOC \& LLM-DOC. 3 comparisons are made between the classification results to examine the effects of including H-PP in classification. First, results from (i) and (ii) are compared to show H-PP’s effects while classification is done with LLM-DOC. Second, results from (iii) and (iv) are compared to show H-PP’s effects under recursive paraphrasing. Lastly, results from (v) are compared to results from (iii) and (iv) to examine the effects of having a full set of human and LLM-generated data. 


In our experiments, we observe that in all 3 sets of comparisons, including H-PP in the classification is effective in promoting TPR@1\%FPR, while its effects on AUROC and accuracy are highly dependent on the presence of watermarking and the type of paraphraser. In the 1st set of comparison, the results show that TPR@1\%FPR increases in all scenarios, but AUROC and accuracy decrease if non-watermarked LLM-DOC are used. For the 2nd set of comparison, AUROC and TPR@1\%FPR increases to a small extent in all scenarios, while accuracy remains unchanged under recursive paraphrasing. Lastly, for the 3rd set of comparison with the full set of data, results vary in 2 extremes depending on the paraphraser used to generate paraphrases from watermarked LLM-DOC, while TPR@1\%FPR increase significantly and AUROC and accuracy decrease slightly with non-watermarked LLM-DOC and their paraphrases. Therefore, it can be concluded that the inclusion of H-PP in classification promotes TPR@1\%FPR with a possible trade- off of AUROC and accuracy. 

Our study has potential to be further extended in the future by studying additional datasets and detection models, to tackle some of the limitations of our study. First, the sentences in the chosen datasets are relatively short, with a mean length of 48.68 tokens. Since the performance of LLM-generated text detectors increases with the input text length, consideration of additional datasets with longer sentences would help provide a more diverse analysis of the effects of H-PP’s inclusion in classification. However, due to the limited availability of datasets that contain H-PP, only datasets with short sentences are used in this project. Second, other state-of-the-art LLM text detection tools could be tested to broaden the findings, such as GPTZero,\footnote{\url{https://gptzero.me/}} which was excluded from our study due to the associated costs.

\bibliographystyle{cas-model2-names}

\bibliography{cas-refs}


\appendix

\section{Examples of LLM-generated Paraphrases}
\label{app:llm-examples}

\begin{longtable}{p{2cm}|m{2cm}|c|p{10.2cm}}
    \caption{Examples of LLM-generated paraphrases; ppi means the i-th round of paraphrase.} \\
    \multicolumn{4}{c}{MRPC} \\
    \hline 
    \textbf{Paraphraser} & \textbf{Input} & \textbf{ppi} & \textbf{Text} \\ 
    \hline
    \multirow{29}{1.5cm}{DIPPER} & \multirow{8}{2cm}{Watermarked GPT Output} & i=0 & The dollar rose 0. 2 percent to \$1.1234 from \$1.1218, after touching a high of \$1.1218 on Friday.\\ 
    && i=1 & The dollar rose 0.10 percent to \$1.1234 from \$1.1218, after a high of \$1.1218 on Friday \\ 
    && i=2 & The dollar rose 0.10 percent to \$1.1234 from \$1.1218. \\ 
    && i=3 & The dollar rose by a penny to \$1.1234 from \$1.1218. \\ 
    && i=4 & The dollar jumped a penny to \$1.1234 from \$1.1218. \\ 
    && i=5 & A little more, the dollar was up a penny to \$1.1234. \\ 
    \cline{2-4}

    & \multirow{11}{2cm}{Watermarked OPT Output} & i=0 & The dollar rose 0.1 percent against the yen to 110.65 yen, after hitting a three-month high of 110.78 yen on Tuesday.\\ 
    && i=1 & The dollar slipped a quarter of a percent against the yen, after hitting a three-month high of 110.65 yen on Tuesday \\ 
    && i=2 & The yen slipped a quarter of a percent against the dollar, after hitting a three-month high of 110.60. \\ 
    && i=3 & It's a good thing that it's not going to be long. The yen is a little less than one yen, the \\ 
    && i=4 & But the yen is a little less than one yen. The yen is just a tenth of a yen. \\ 
    && i=5 & This is the amount of ten thousand yen, but the yen is only ten thousand yen. \\  
    
    \cline{2-4}
    & \multirow{8}{2cm}{Non-watermarked GPT Output} & i=0 & The dollar rose 0.3 percent to \$1.3240 per euro, while the euro was little changed at \$1.3240. The dollar index\\ 
    && i=1 & The dollar was gaining 0.3 percent to \$1.3240, and the Euro was less than \$1,400. \\ 
    && i=2 & The dollar was up 0.3 percent to \$1.3240. The Euro was less than \$1,400. \\ 
    && i=3 & It was a little over \$2,000. The Euro was not much more than \$3,000. \\ 
    && i=4 & The Euro was not much more than a thousand roubles. \\ 
    && i=5 & The Euro was less than a thousand rubles. \\  

    \cline{2-4}

    & \multirow{12}{2cm}{Non-watermarked OPT Output} & i=0 & The dollar rose 0.2 percent against the yen to 111.85 yen, after hitting a two-week high of 111.90 yen on Friday.\\ 
    && i=1 & On Friday the yen rose by 0.2 percent against the yen. The yen fell by 0.2 percent against the dollar. The pair had hit \\ 
    && i=2 & The pair had fallen to the tenth day of the previous month. On Friday the yen rose by 0.2 percent against the dollar. \\ 
    && i=3 & On Friday the yen rose by 0.2 percent against the dollar. The pair had fallen to the tenth day of the previous month. \\ 
    && i=4 & But if you were to use the yen as an example, you would have a hard time keeping up the momentum. On Friday, the dollar fell by \\ 
    && i=5 & But you would have to think of the yen as an example, and I doubt if you would be able to sustain the momentum. \\  
    \hline
    \multirow{48}{1.5cm}{BART} & \multirow{12}{2cm}{Watermarked GPT Output} & i=0 & The dollar rose 0. 2 percent to \$1.1234 from \$1.1218, after touching a high of \$1.1218 on Friday.\\ 
    && i=1 & The dollar rose 0.2 percent to \$1.1234 on the New York Stock Exchange, after touching a high of \$.1218 on Friday. \\ 
    && i=2 & The dollar rose 0.2 percent to \$1.1234 on the New York Stock Exchange, after touching a high of \$.1218 on Friday. \\ 
    && i=3 & The dollar rose 0.2 percent to \$1.1234 on the New York Stock Exchange, after touching a high of \$.1218 on Friday. \\ 
    && i=4 & The dollar rose 0.2 percent to \$1.1234 on the New York Stock Exchange, after touching a high of \$.1218 on Friday. \\ 
    && i=5 & The dollar rose 0.2 percent to \$1.1234 on the New York Stock Exchange, after touching a high of \$.1218 on Friday. \\  
    \cline{2-4}
    
    & \multirow{12}{2cm}{Watermarked OPT Output} & i=0 & The dollar rose 0.1 percent against the yen to 110.65 yen, after hitting a three-month high of 110.78 yen on Tuesday.\\ 
    && i=1 & The dollar rose 0.1 percent against the yen to 110.65 yen, after hitting a three-month high of 110.78 yen on Tuesday. \\ 
    && i=2 & The dollar rose 0.1 percent against the yen to 110.65 yen, after hitting a three-month high of 110.78 yen on Tuesday. \\ 
    && i=3 & The dollar rose 0.1 percent against the yen to 110.65 yen, after hitting a three-month high of 110.78 yen on Tuesday. \\ 
    && i=4 & The dollar rose 0.1 percent against the yen to 110.65 yen, after hitting a three-month high of 110.78 yen on Tuesday. \\ 
    && i=5 & The dollar rose 0.1 percent against the yen to 110.65 yen, after hitting a three-month high of 110.78 yen on Tuesday. \\  

    \cline{2-4}
    & \multirow{12}{2cm}{Non-watermarked GPT Output} & i=0 & The dollar rose 0.3 percent to \$1.3240 per euro, while the euro was little changed at \$1.3240. The dollar index\\ 
    && i=1 & The dollar rose 0.3 percent against the euro to \$1.3240, while the euro was little changed against the dollar at 1.3243. \\ 
    && i=2 & The dollar rose 0.3 percent against the euro to \$1.3240, while the euro was little changed against the dollar at 1.3243. \\ 
    && i=3 & The dollar rose 0.3 percent against the euro to \$1.3240, while the euro was little changed against the dollar at 1.3243. \\ 
    && i=4 & The dollar rose 0.3 percent against the euro to \$1.3240, while the euro was little changed against the dollar at 1.3243. \\ 
    && i=5 & The dollar rose 0.3 percent against the euro to \$1.3240, while the euro was little changed against the dollar at 1.3243. \\  

    \cline{2-4}
    & \multirow{12}{2cm}{Non-watermarked OPT Output} & i=0 & The dollar rose 0.2 percent against the yen to 111.85 yen, after hitting a two-week high of 111.90 yen on Friday.\\ 
    && i=1 & The dollar rose 0.2 percent against the yen to 110.85 yen, after hitting a two-week high of 110.90 yen on Friday. \\ 
    && i=2 & The dollar rose 0.2 percent against the yen to 110.85 yen, after hitting a two-week high of 110.90 yen on Friday. \\ 
    && i=3 & The dollar rose 0.2 percent against the yen to 110.85 yen, after hitting a two-week high of 110.90 yen on Friday. \\ 
    && i=4 & The dollar rose 0.2 percent against the yen to 110.85 yen, after hitting a two-week high of 110.90 yen on Friday. \\ 
    && i=5 & The dollar rose 0.2 percent against the yen to 110.85 yen, after hitting a two-week high of 110.90 yen on Friday. \\
\end{longtable}

\section{Classification Results with Human-written Document/ Paraphrases and LLM-generated Paraphrases}
\label{appendix a.3}

\begin{figure}[!htb]
    \caption{ROC Curve from OpenAI Detector with Human-generated Documents / Paraphrases and Non-watermarked DIPPER (left) / BART-generated (right) Paraphrases from MRPC OPT-Generated Text}
	\centering
	\includegraphics[scale=.32]{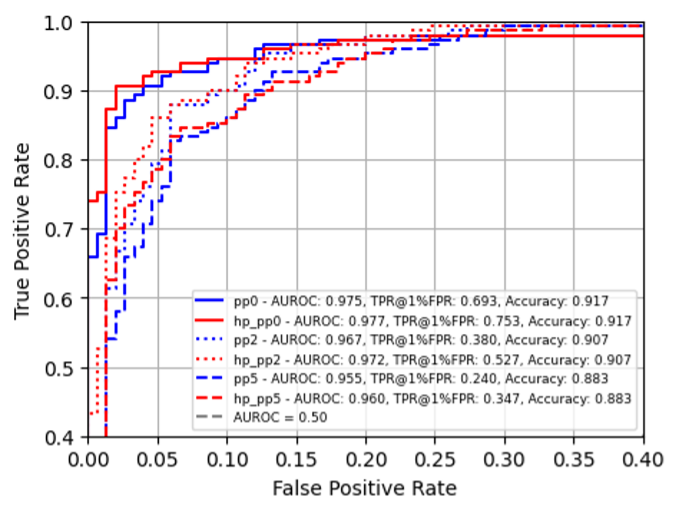}
    \includegraphics[scale=.32]{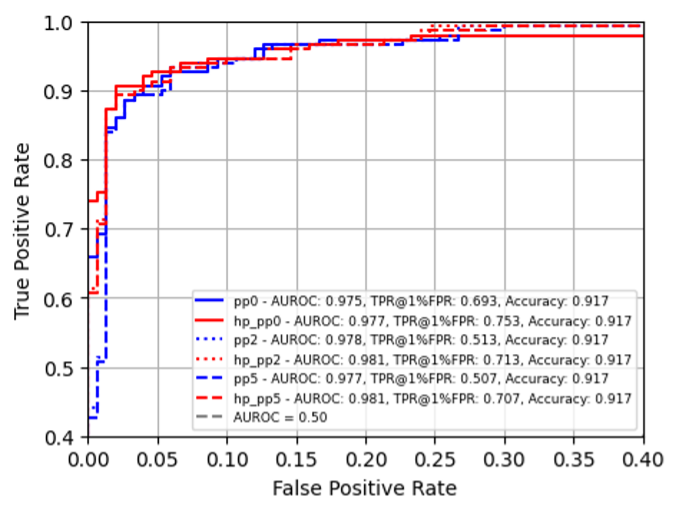}
\end{figure}

\begin{figure}[!htb]
    \caption{ROC Curve from Watermark Detector with Human-generated Documents / Paraphrases and Watermarked DIPPER (left) / BART-generated (right) Paraphrases from MRPC OPT-Generated Text}
	\centering
	\includegraphics[scale=.32]{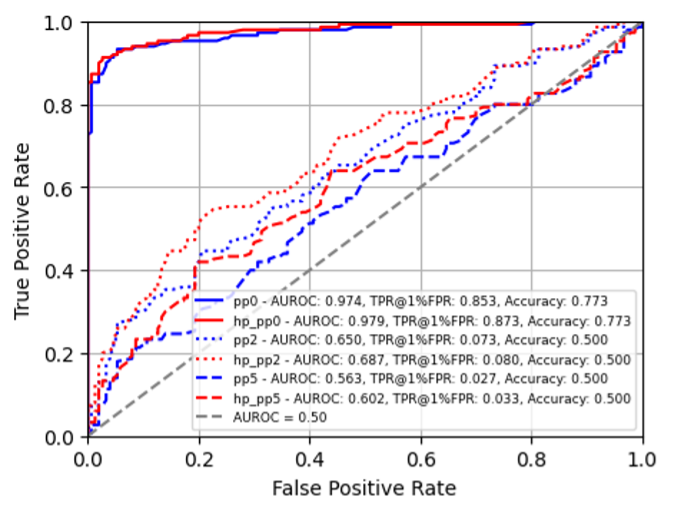}
    \includegraphics[scale=.32]{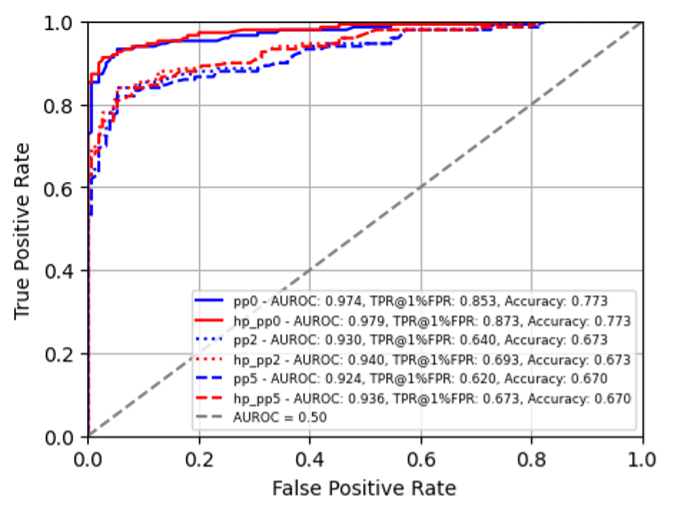}
\end{figure}

\begin{figure}[!htb]
    \caption{ROC Curve from OpenAI Detector with Human-generated Documents / Paraphrases and Non-watermarked DIPPER (left) / BART-generated (right) Paraphrases from XSum GPT-Generated Text}
	\centering
	\includegraphics[scale=.32]{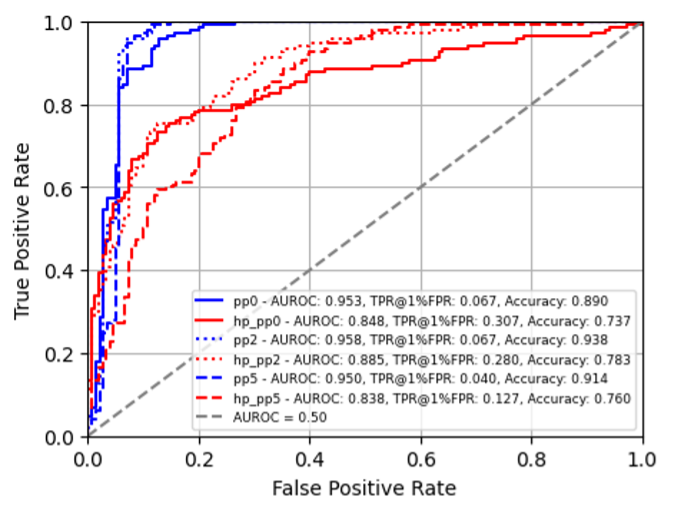}
    \includegraphics[scale=.32]{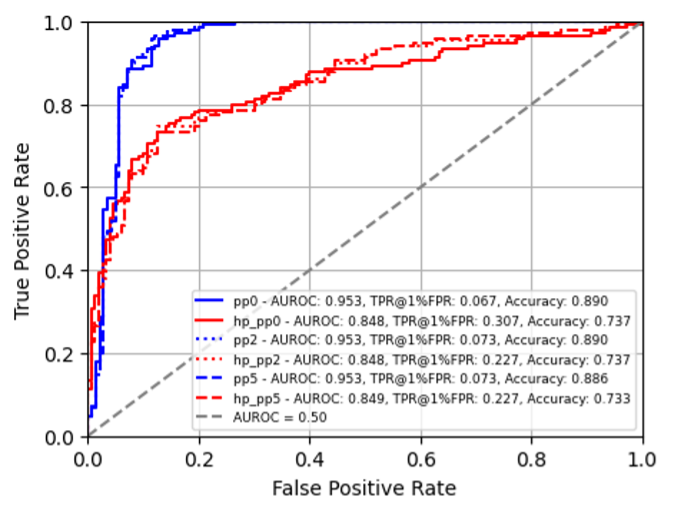}
\end{figure}

\begin{figure}[!htb]
    \caption{ROC Curve from Watermark Detector with Human-generated Documents / Paraphrases and Watermarked DIPPER (left) / BART-generated (right) Paraphrases from XSum GPT-Generated Text}
	\centering
	\includegraphics[scale=.32]{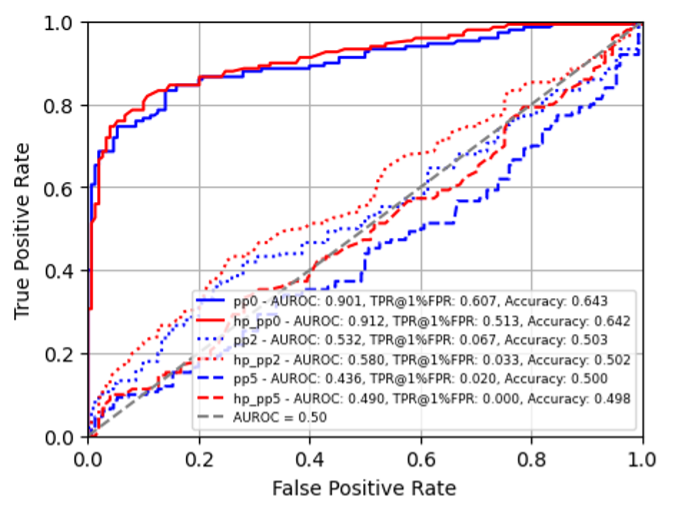}
    \includegraphics[scale=.32]{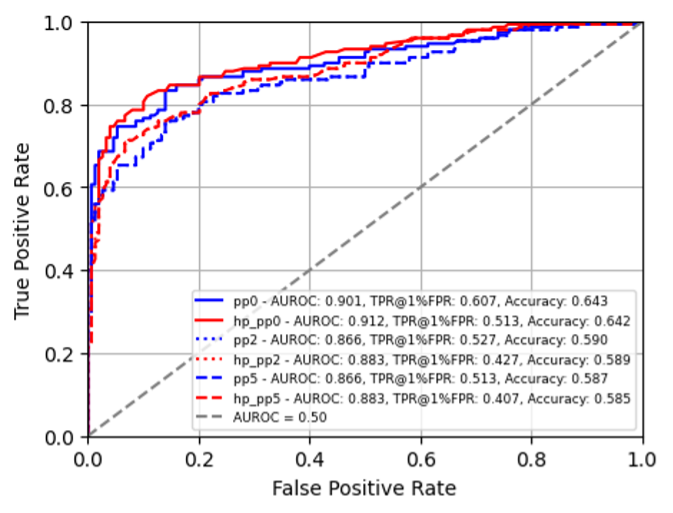}
\end{figure}

\begin{figure}[!htb]
    \caption{ROC Curve from OpenAI Detector with Human-generated Documents / Paraphrases and Non-watermarked DIPPER (left) / BART-generated (right) Paraphrases from XSum OPT-Generated Text}
	\centering
	\includegraphics[scale=.32]{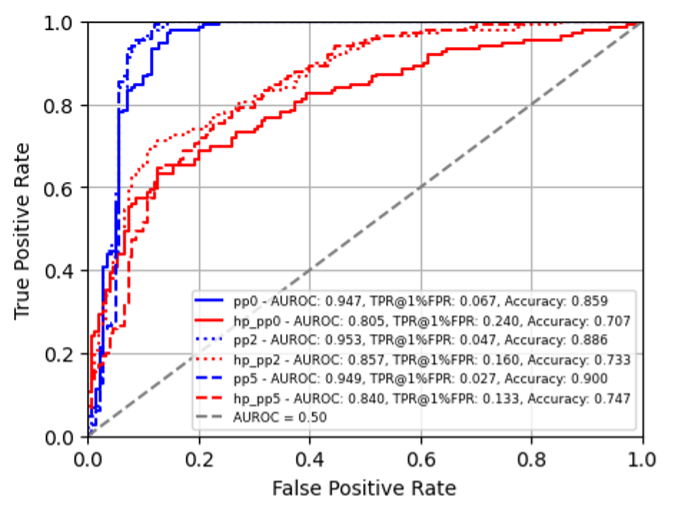}
    \includegraphics[scale=.32]{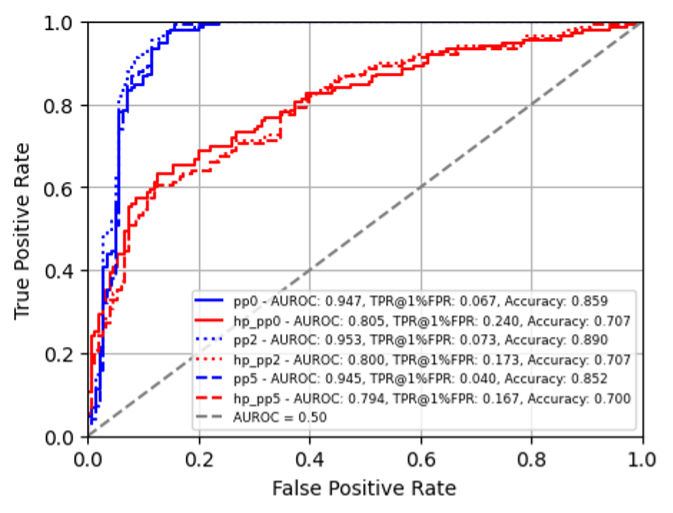}
\end{figure}

\begin{figure}[!htb]
    \caption{ROC Curve from Watermark Detector with Human-generated Documents / Paraphrases and Watermarked DIPPER (left) / BART-generated (right) Paraphrases from XSum OPT-Generated Text}
	\centering
	\includegraphics[scale=.32]{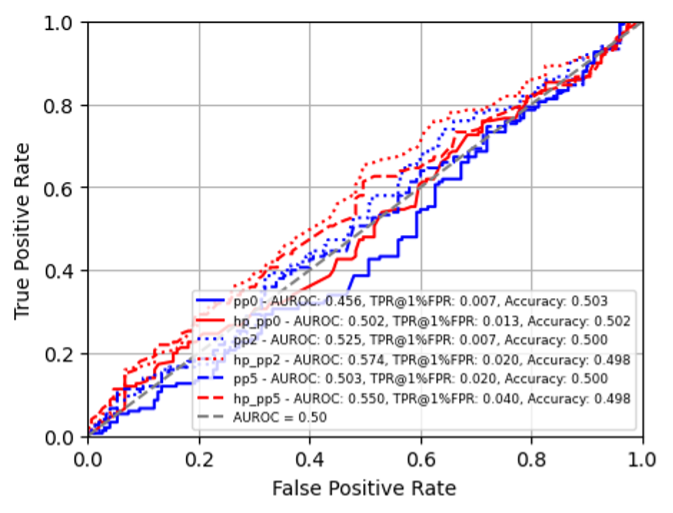}
    \includegraphics[scale=.32]{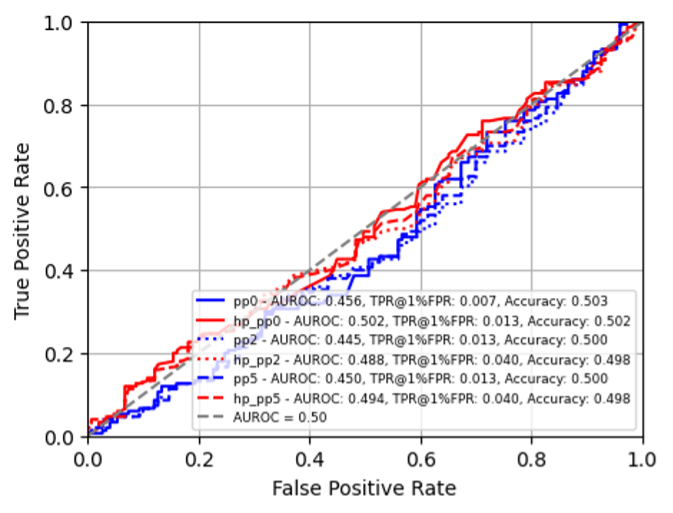}
\end{figure}

\begin{figure}[!htb]
    \caption{ROC Curve from OpenAI Detector with Human-generated Documents / Paraphrases and Non-watermarked DIPPER (left) / BART-generated (right) Paraphrases from QQP GPT-Generated Text}
	\centering
	\includegraphics[scale=.32]{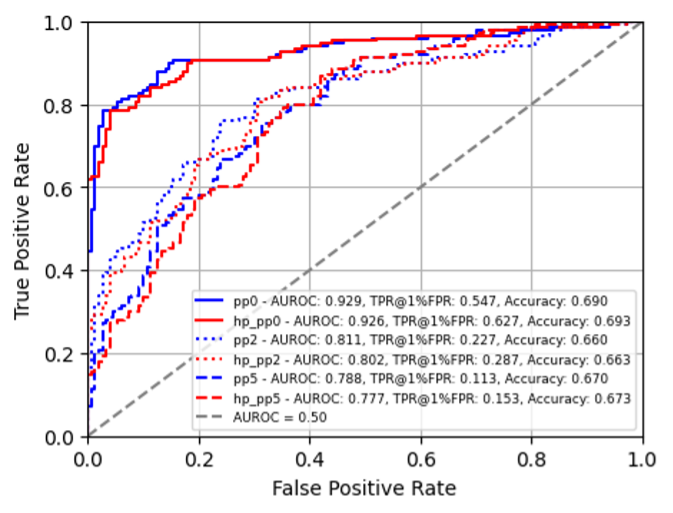}
    \includegraphics[scale=.32]{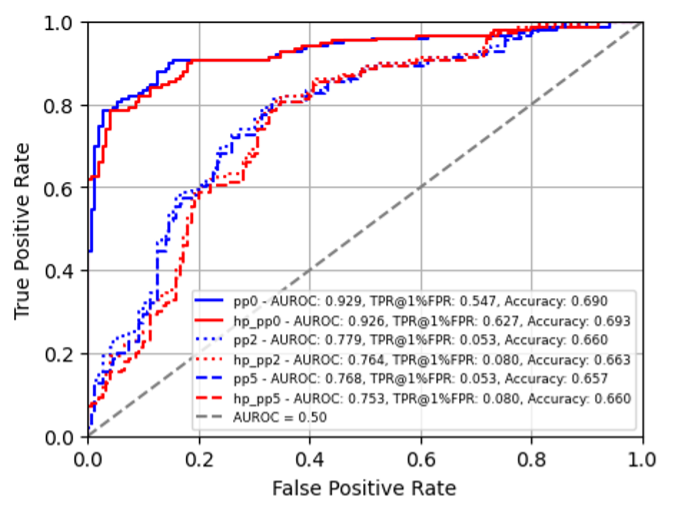}
\end{figure}

\begin{figure}[!htb]
    \caption{ROC Curve from Watermark Detector with Human-generated Documents / Paraphrases and Watermarked DIPPER (left) / BART-generated (right) Paraphrases from QQP GPT-Generated Textt}
	\centering
	\includegraphics[scale=.32]{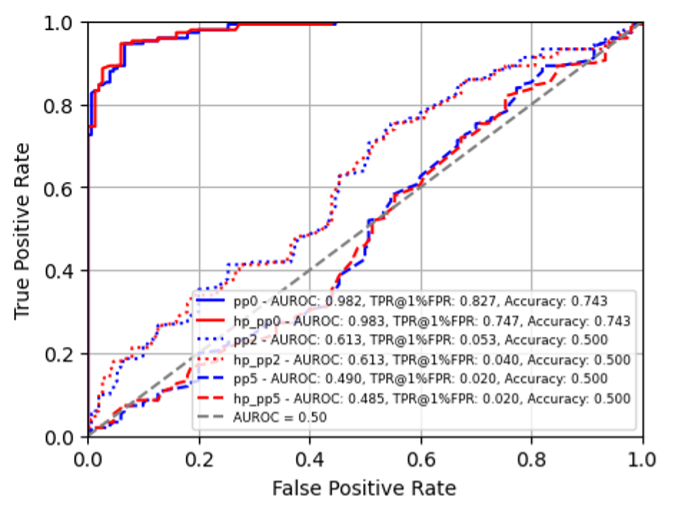}
    \includegraphics[scale=.32]{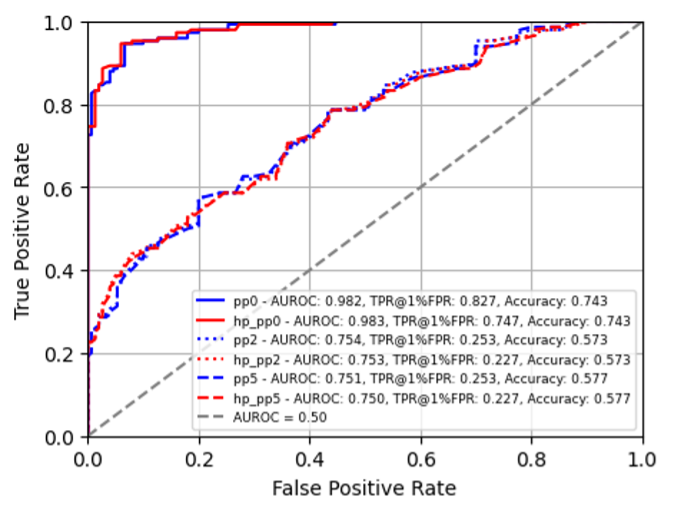}
\end{figure}

\begin{figure}[!htb]
    \caption{ROC Curve from OpenAI Detector with Human-generated Documents / Paraphrases and Non-watermarked DIPPER (left) / BART-generated (right) Paraphrases from QQP OPT-Generated Text}
	\centering
	\includegraphics[scale=.32]{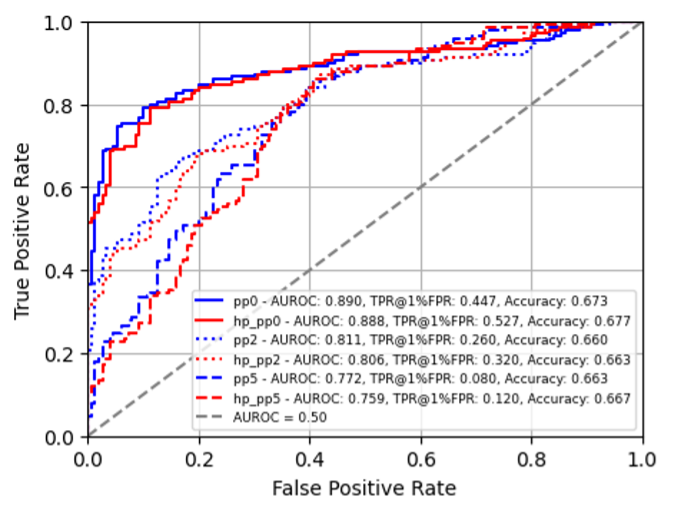}
    \includegraphics[scale=.32]{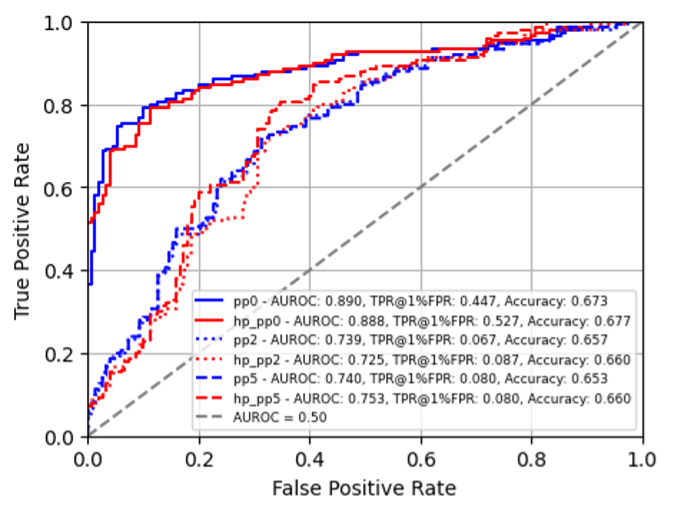}
\end{figure}

\begin{figure}[hbt!]
    \caption{ROC Curve from Watermark Detector with Human-generated Documents / Paraphrases and Watermarked DIPPER (left) / BART-generated (right) Paraphrases from QQP OPT-Generated Text}
	\centering
	\includegraphics[scale=.32]{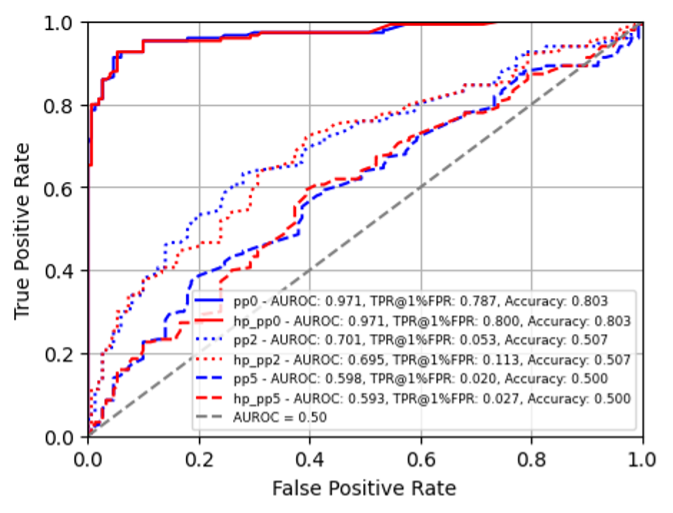}
    \includegraphics[scale=.32]{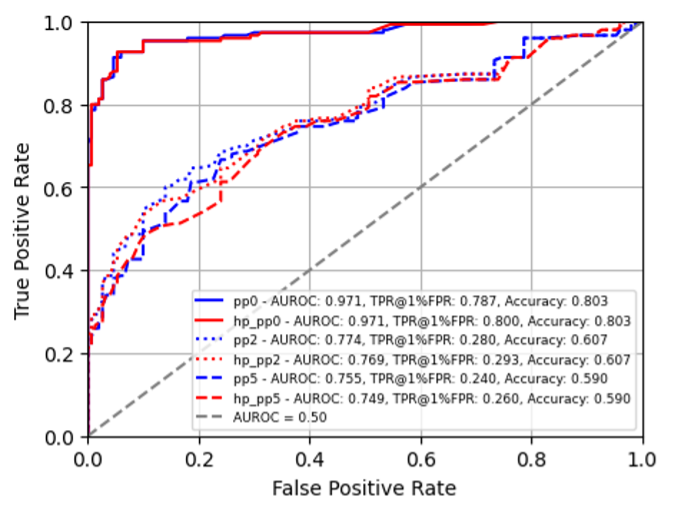}
\end{figure}

\begin{figure}[!htb]
    \caption{ROC Curve from OpenAI Detector with Human-generated Documents / Paraphrases and Non-watermarked DIPPER (left) / BART-generated (right) Paraphrases from MultiPIT GPT-Generated Text}
	\centering
	\includegraphics[scale=.32]{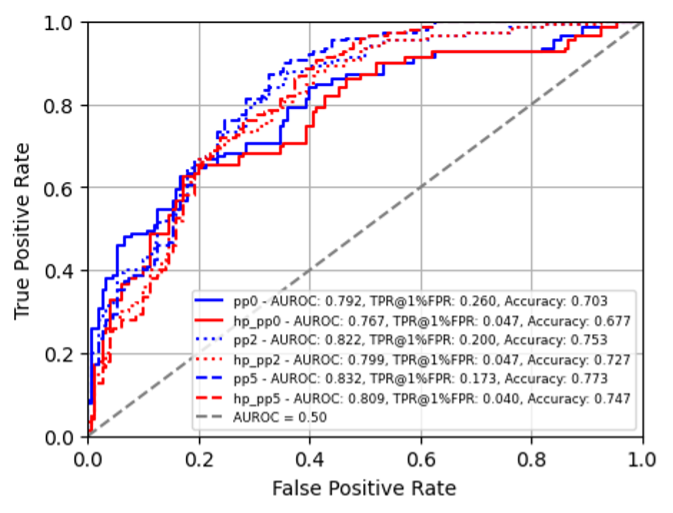}
    \includegraphics[scale=.32]{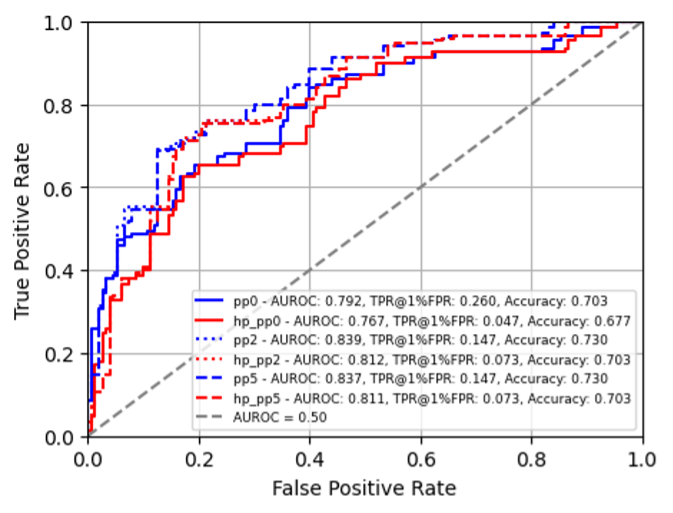}
\end{figure}

\begin{figure}[!htb]
    \caption{ROC Curve from Watermark Detector with Human-generated Documents / Paraphrases and Watermarked DIPPER (left) / BART-generated (right) Paraphrases from MultiPIT GPT-Generated Text}
	\centering
	\includegraphics[scale=.32]{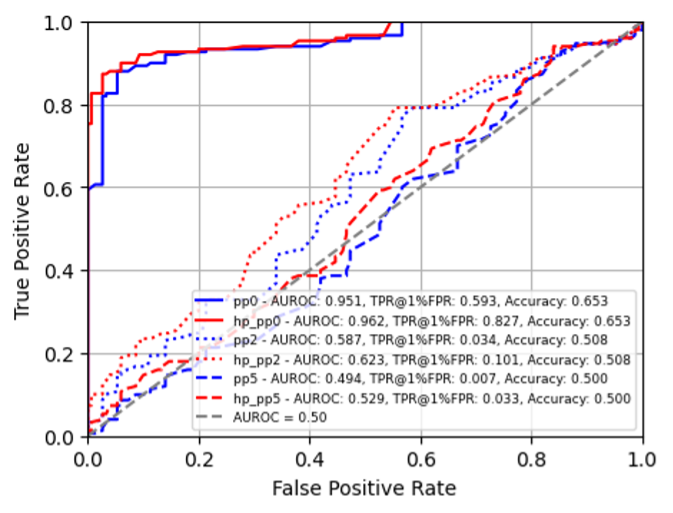}
    \includegraphics[scale=.32]{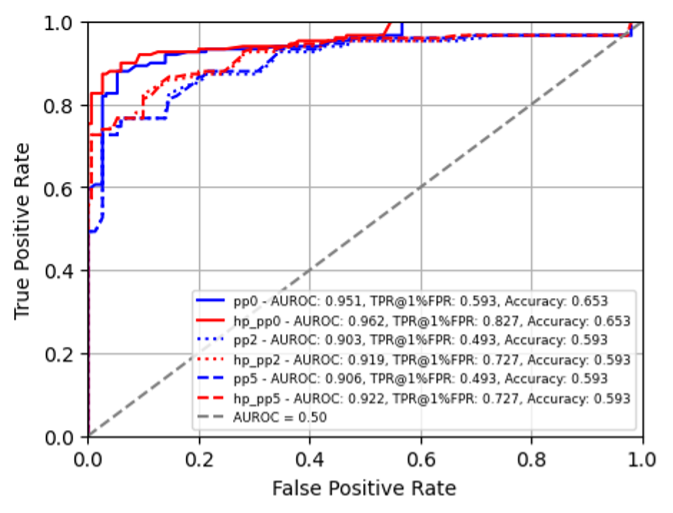}
\end{figure}

\begin{figure}[!htb]
    \caption{ROC Curve from OpenAI Detector with Human-generated Documents / Paraphrases and Non-watermarked DIPPER (left) / BART-generated (right) Paraphrases from MultiPIT OPT-Generated Text}
	\centering
	\includegraphics[scale=.32]{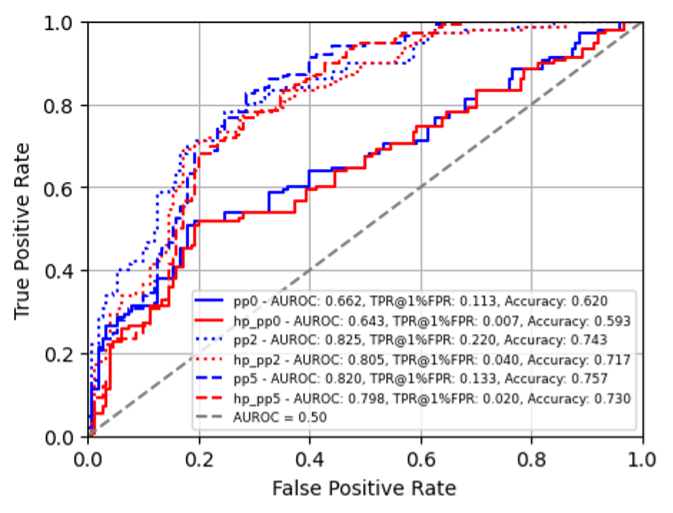}
    \includegraphics[scale=.32]{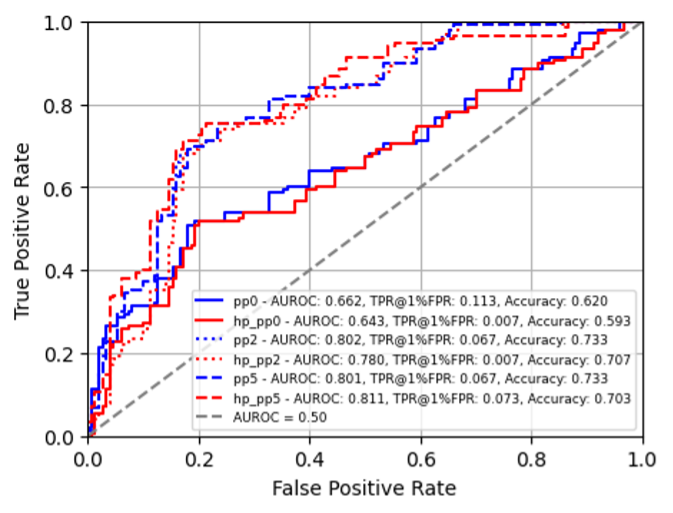}
\end{figure}

\begin{figure}[!htb]
    \caption{ROC Curve from Watermark Detector with Human-generated Documents / Paraphrases and Watermarked DIPPER (left) / BART-generated (right) Paraphrases from MultiPIT OPT-Generated Text}
	\centering
	\includegraphics[scale=.32]{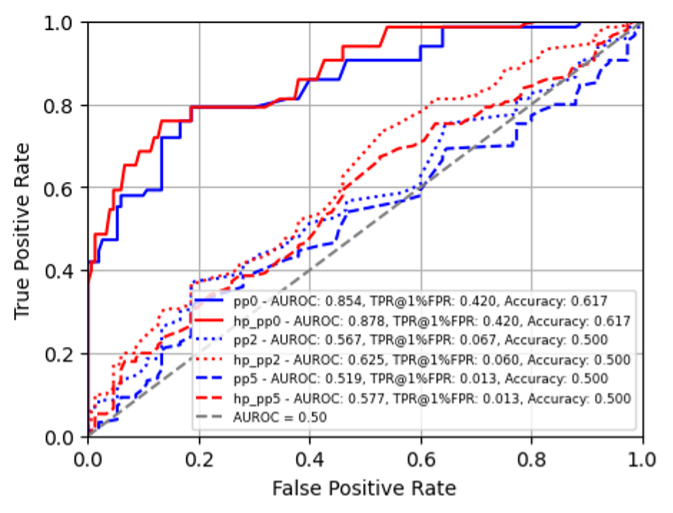}
    \includegraphics[scale=.32]{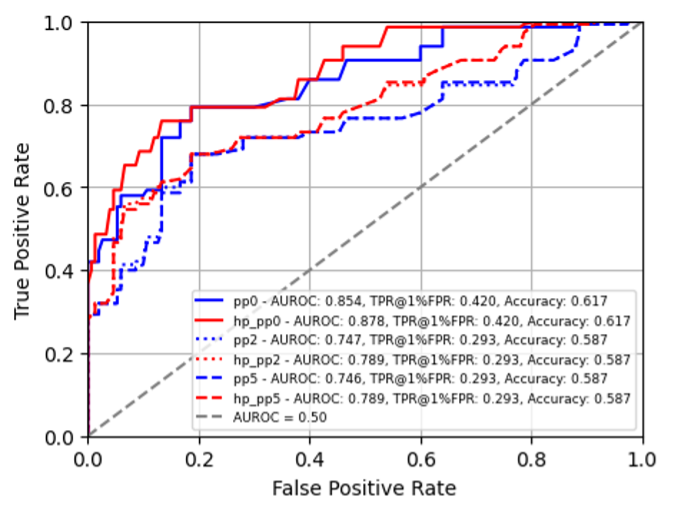}
\end{figure}

\section{Classification Results with Full Human-written Data and LLM-generated Data}
\label{appendix a.4}

\begin{figure}[hbt!]
    \caption{ROC Curve with Full Human Data and Non-watermarked MRPC LLM-Data (Paraphrases generated with DIPPER (left) and BART (right) from MRPC GPT2-generated Text)}
	\centering
	\includegraphics[scale=.32]{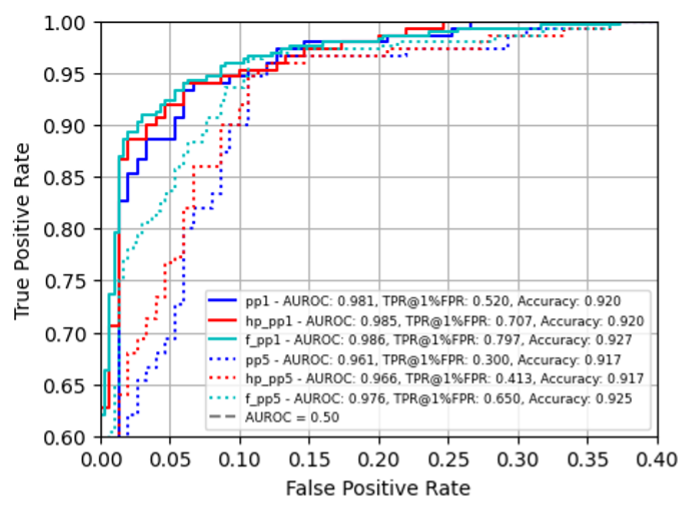}
    \includegraphics[scale=.32]{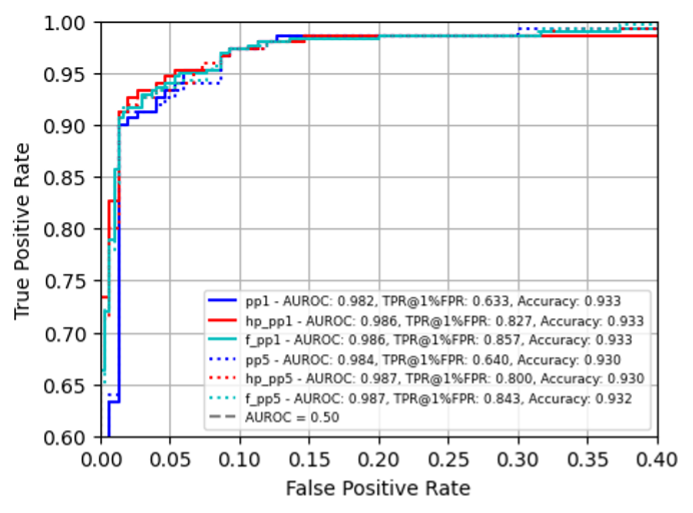}
\end{figure}

\begin{figure}[hbt!]
    \caption{ROC Curve with Full Human Data and Watermarked MRPC LLM -Data (Paraphrases generated with DIPPER (left) and BART (right) from MRPC GPT2-generated Text)}
	\centering
	\includegraphics[scale=.32]{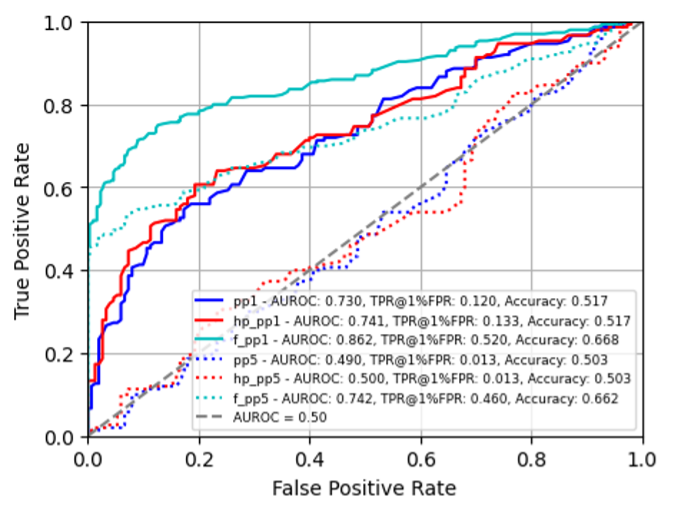}
    \includegraphics[scale=.32]{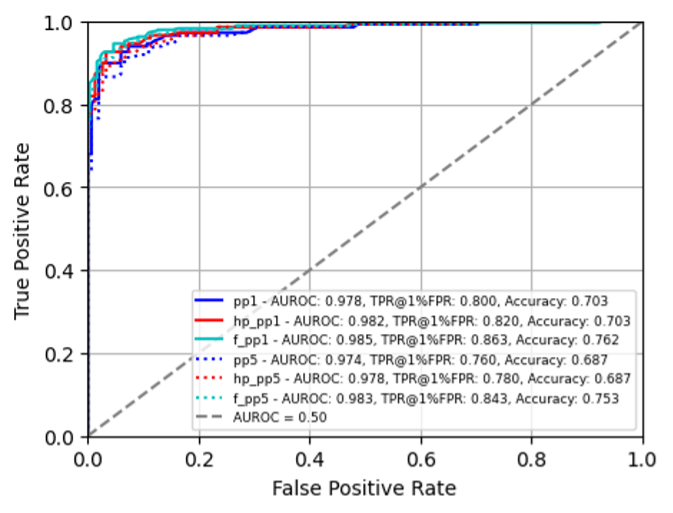}
\end{figure}

\begin{figure}[hbt!]
    \caption{ROC Curve with Full Human Data and Non-watermarked MRPC LLM-Data (Paraphrases generated with DIPPER (left) and BART (right) from MRPC OPT-generated Text)}
	\centering
	\includegraphics[scale=.32]{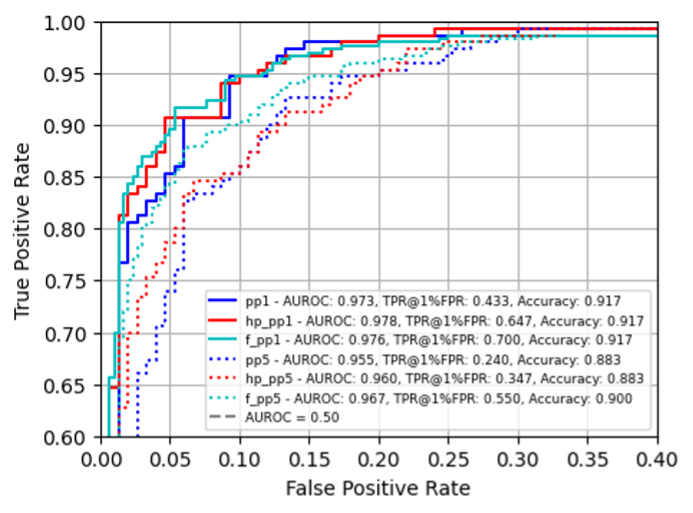}
    \includegraphics[scale=.32]{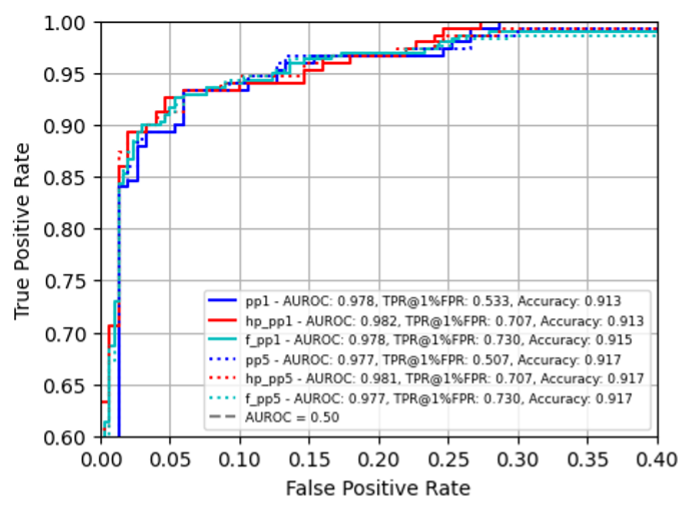}
\end{figure}

\begin{figure}[hbt!]
    \caption{ROC Curve with Full Human Data and Watermarked MRPC LLM-Data (Paraphrases generated with DIPPER (left) and BART (right) from MRPC OPT-generated Text)}
	\centering
	\includegraphics[scale=.32]{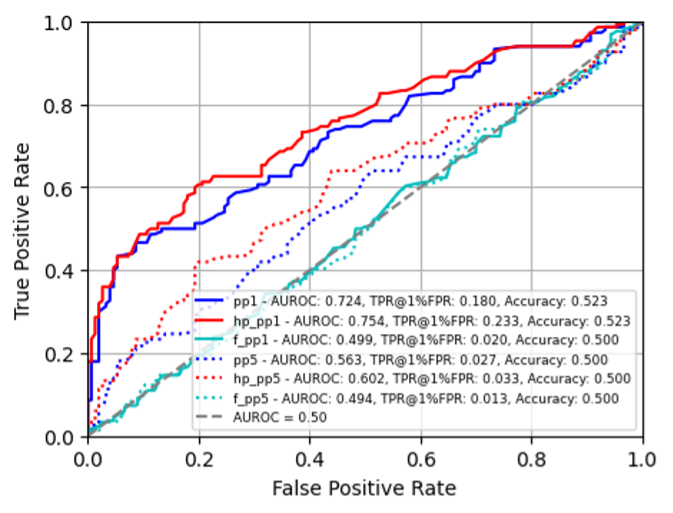}
    \includegraphics[scale=.32]{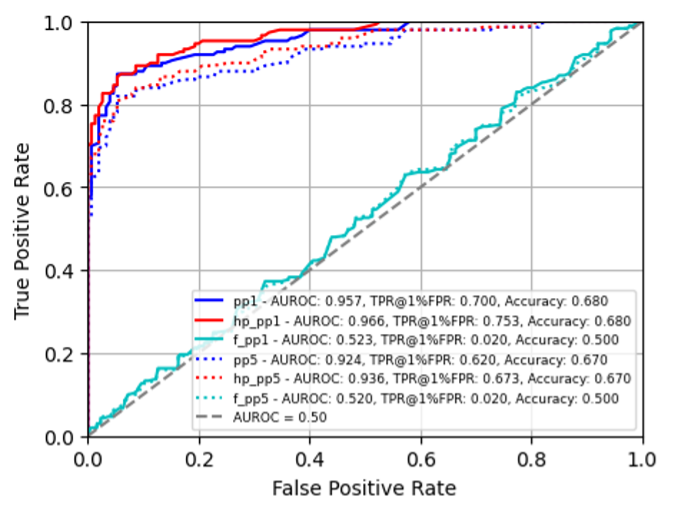}
\end{figure}

\begin{figure}[hbt!]
    \caption{ROC Curve with Full Human Data and Non-watermarked XSum LLM-Data (Paraphrases generated with DIPPER (left) and BART (right) from XSum GPT-generated Text)}
	\centering
	\includegraphics[scale=.32]{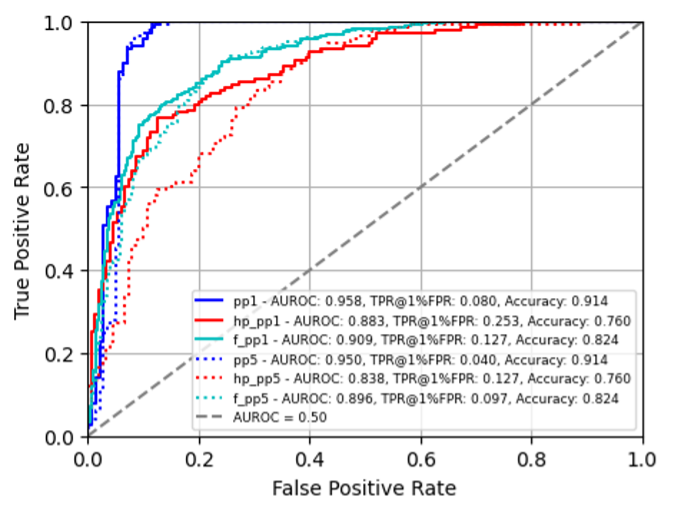}
    \includegraphics[scale=.32]{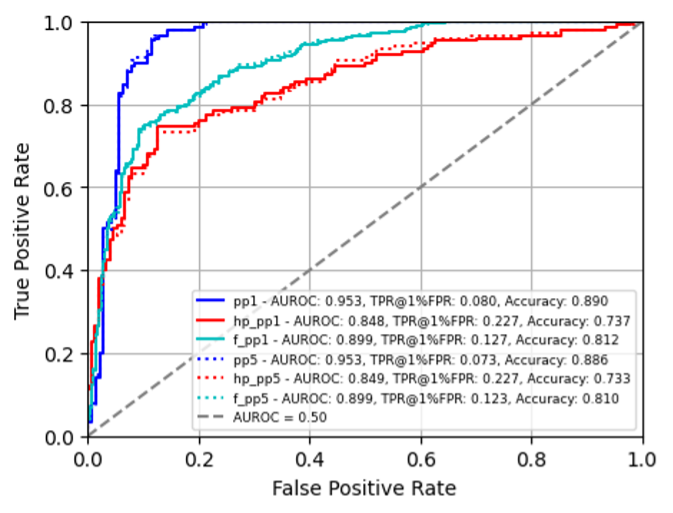}
\end{figure}

\begin{figure}[hbt!]
    \caption{ROC Curve with Full Human Data and Watermarked XSum LLM-Data (Paraphrases generated with DIPPER (left) and BART (right) from XSum GPT-generated Text)}
	\centering
	\includegraphics[scale=.32]{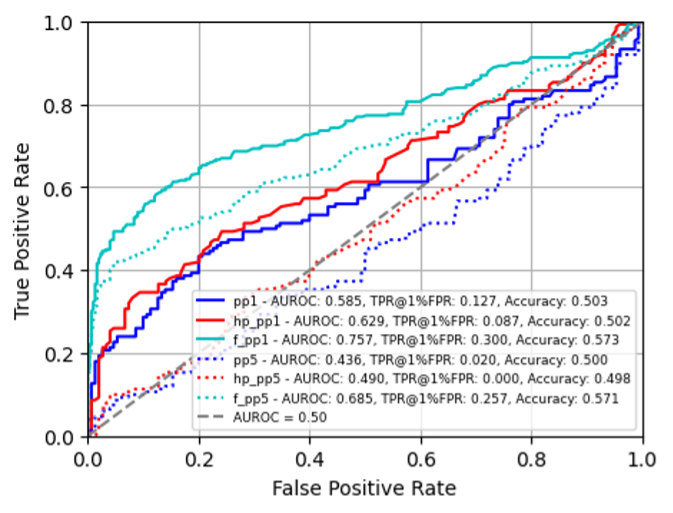}
    \includegraphics[scale=.32]{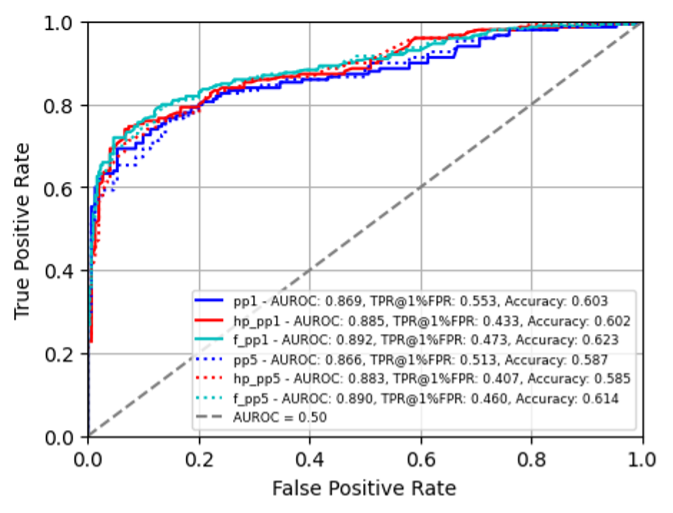}
\end{figure}

\begin{figure}[hbt!]
    \caption{ROC Curve with Full Human Data and Non-watermarked XSum LLM-Data (Paraphrases generated with DIPPER (left) and BART (right) from XSum OPT-generated Text)}
	\centering
	\includegraphics[scale=.32]{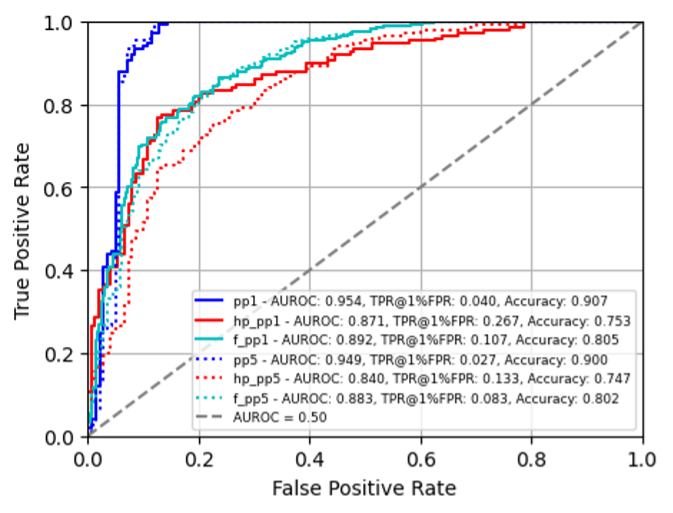}
    \includegraphics[scale=.32]{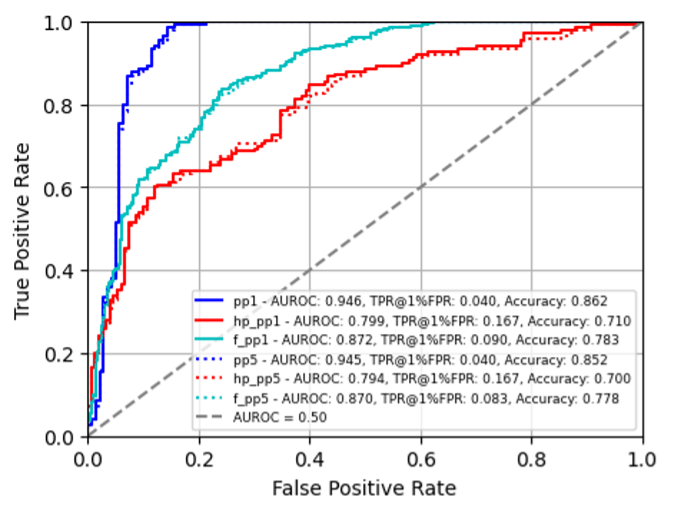}
\end{figure}

\begin{figure}[hbt!]
    \caption{ROC Curve with Full Human Data and Watermarked XSum LLM-Data (Paraphrases generated with DIPPER (left) and BART (right) from XSum OPT-generated Text)}
	\centering
	\includegraphics[scale=.32]{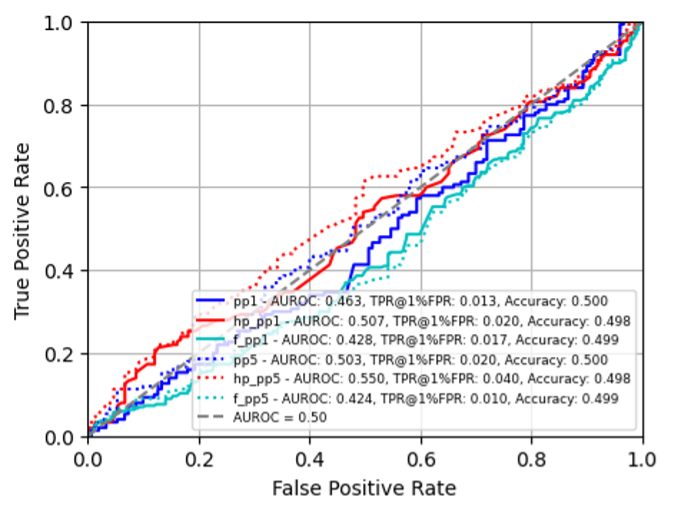}
    \includegraphics[scale=.32]{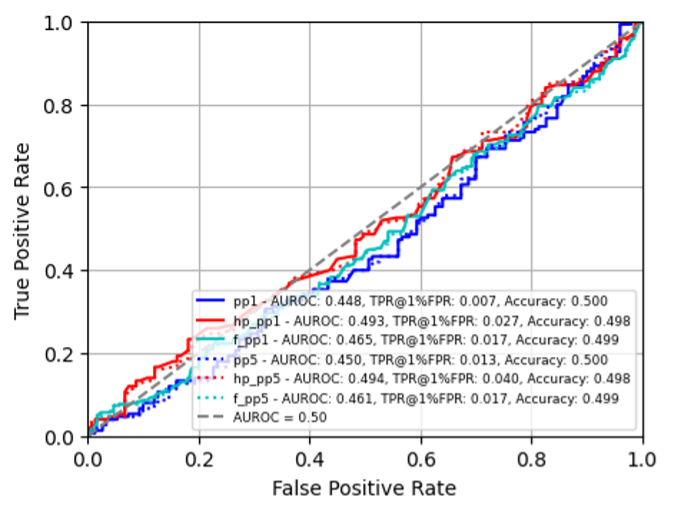}
\end{figure}

\begin{figure}[hbt!]
    \caption{ROC Curve with Full Human Data and Non-watermarked QQP LLM-Data (Paraphrases generated with DIPPER (left) and BART (right) from QQP GPT-generated Text)}
	\centering
	\includegraphics[scale=.32]{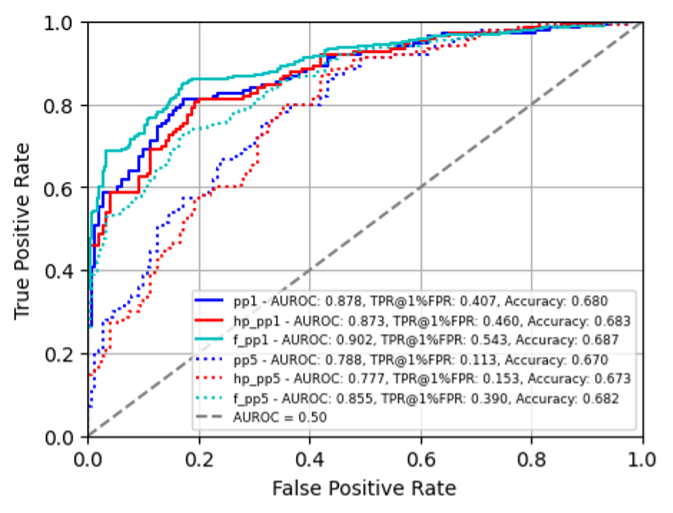}
    \includegraphics[scale=.32]{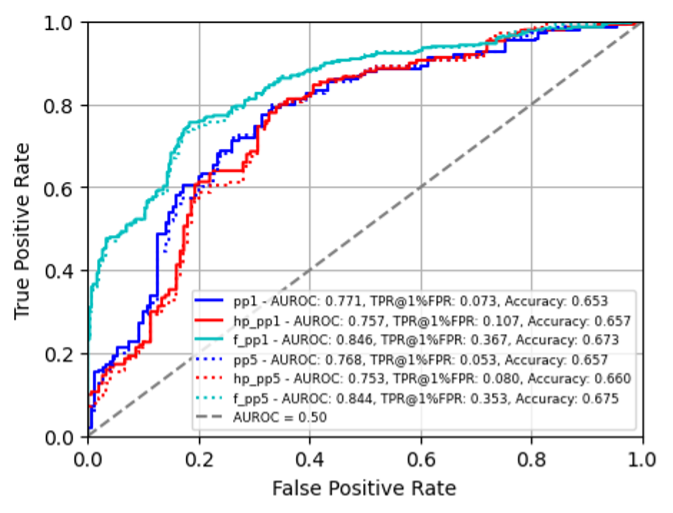}
\end{figure}

\begin{figure}[hbt!]
\caption{ROC Curve with Full Human Data and Watermarked QQP LLM-Data (Paraphrases generated with DIPPER (left) and BART (right) from QQP GPT-generated Text)}
	\centering
	\includegraphics[scale=.32]{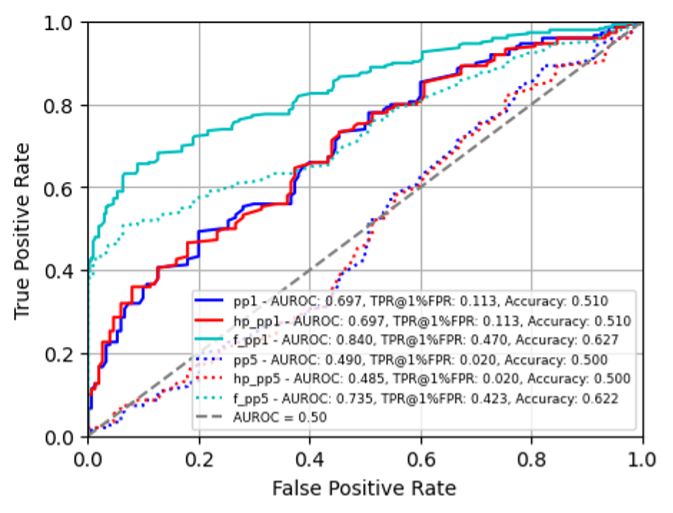}
    \includegraphics[scale=.32]{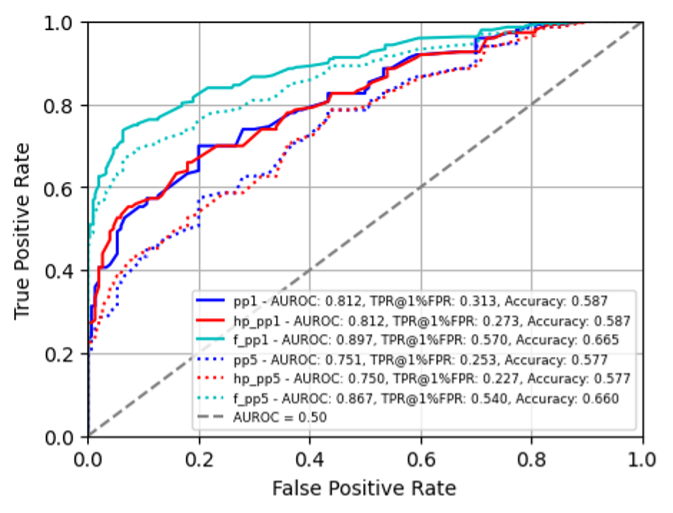}
\end{figure}

\begin{figure}[hbt!]
    \caption{ROC Curve with Full Human Data and Non-watermarked QQP LLM-Data (Paraphrases generated with DIPPER (left) and BART (right) from QQP OPT-generated Text)}
	\centering
	\includegraphics[scale=.32]{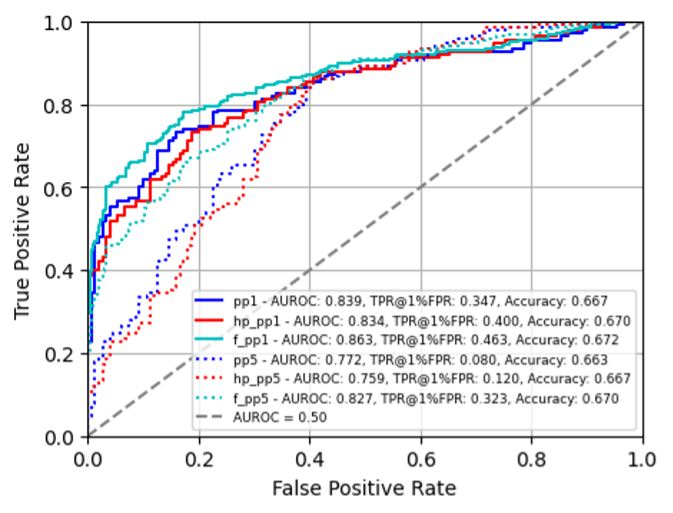}
    \includegraphics[scale=.32]{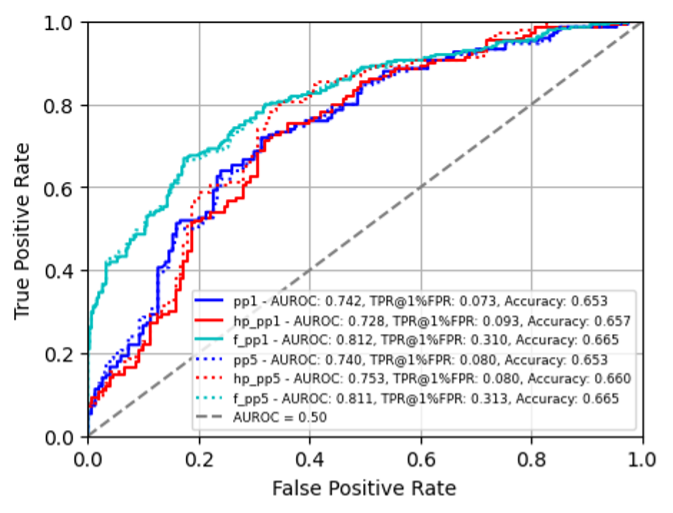}
\end{figure}

\begin{figure}[hbt!]
    \caption{ROC Curve with Full Human Data and Watermarked QQP LLM-Data (Paraphrases generated with DIPPER (left) and BART (right) from QQP OPT-generated Text)}
	\centering
	\includegraphics[scale=.32]{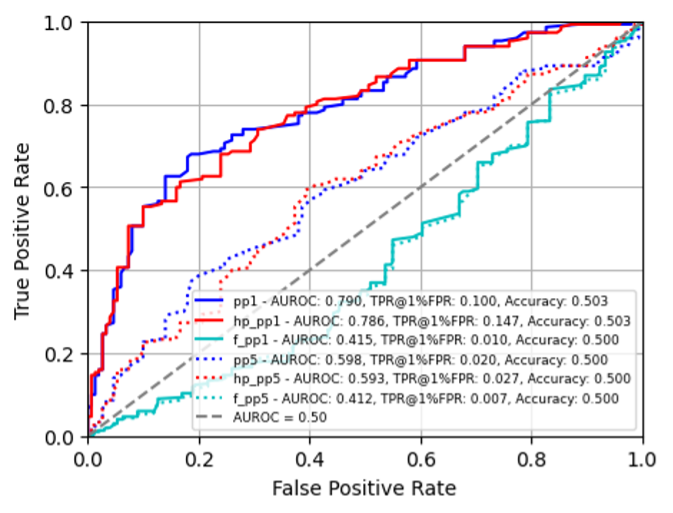}
    \includegraphics[scale=.32]{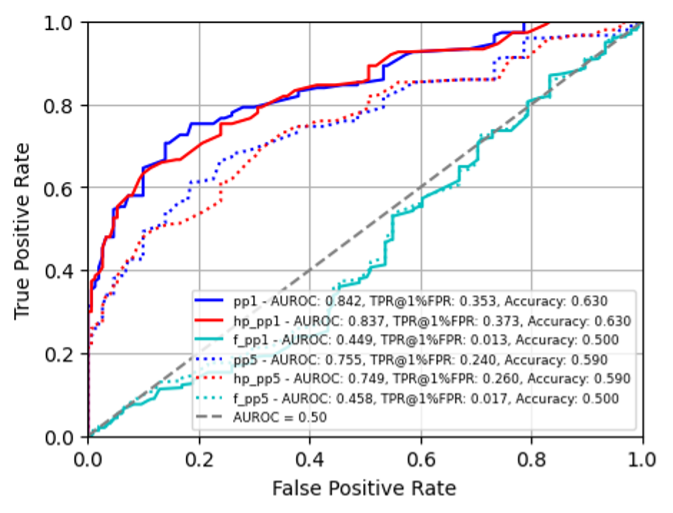}
\end{figure}

\begin{figure}[hbt!]
    \caption{ROC Curve with Full Human Data and Non-watermarked MultiPIT LLM-Data (Paraphrases generated with DIPPER (left) and MultiPIT (right) from MultiPIT GPT-generated Text)}
	\centering
    \includegraphics[scale=.32]{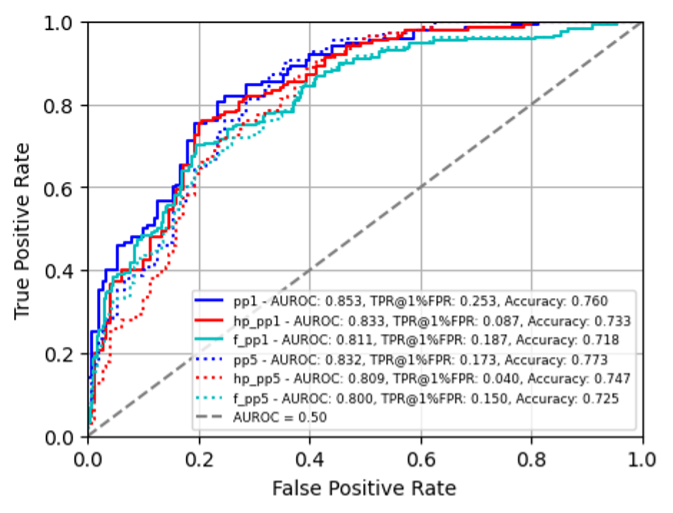}
    \includegraphics[scale=.32]{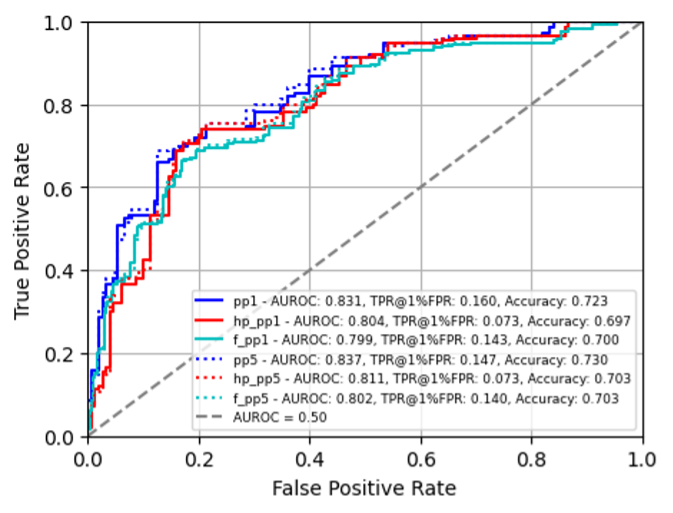}
\end{figure}

\begin{figure}[hbt!]
    \caption{ROC Curve with Full Human Data and Watermarked MultiPIT LLM-Data (Paraphrases generated with DIPPER (left) and BART (right) from MultiPIT GPT-generated Text)}
	\centering
	\includegraphics[scale=.32]{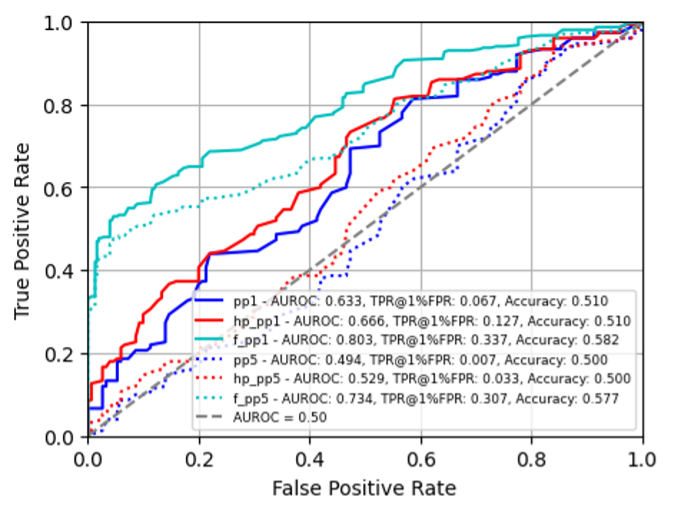}
    \includegraphics[scale=.32]{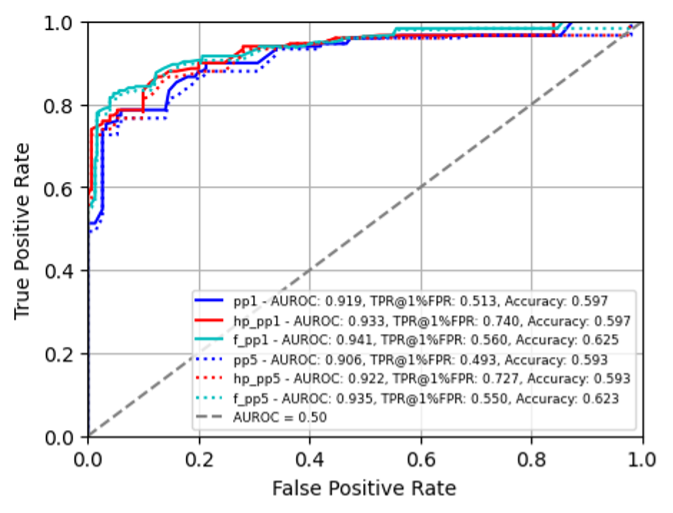}
\end{figure}

\begin{figure}[hbt!]
    \caption{ROC Curve with Full Human Data and Non-watermarked MultiPIT LLM-Data (Paraphrases generated with DIPPER (left) and BART (right) from MultiPIT OPT-generated Text)}
	\centering
	\includegraphics[scale=.32]{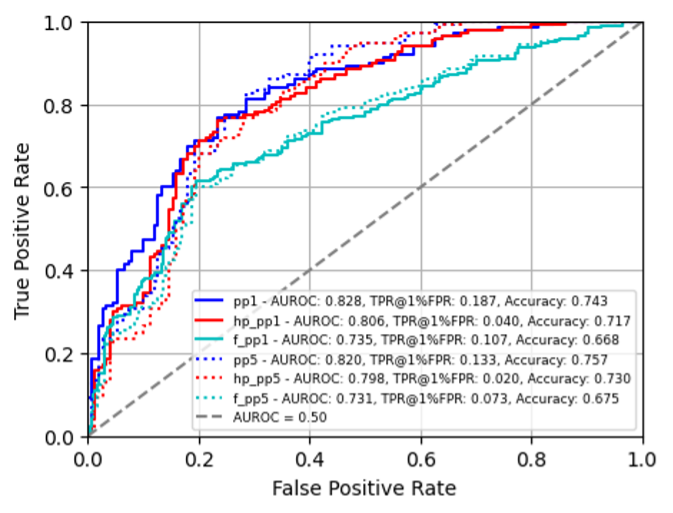}
    \includegraphics[scale=.32]{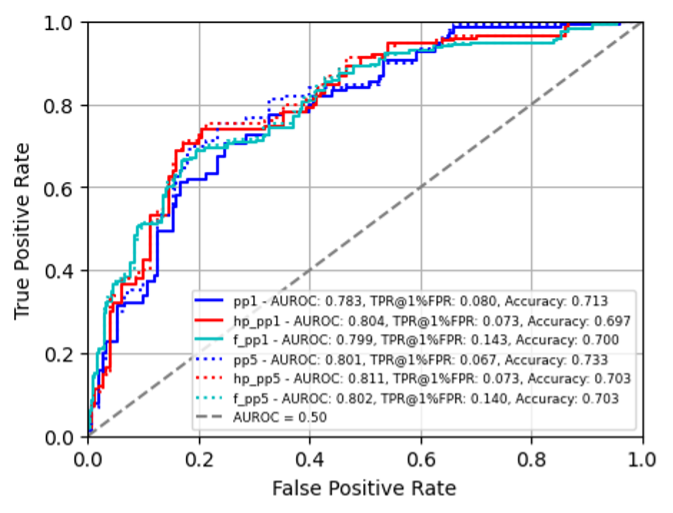}
\end{figure}

\begin{figure}[hbt!]
    \caption{ROC Curve with Full Human Data and Watermarked MultiPIT LLM-Data (Paraphrases generated with DIPPER (left) and BART (right) from MultiPIT OPT-generated Text)}
	\centering
	\includegraphics[scale=.32]{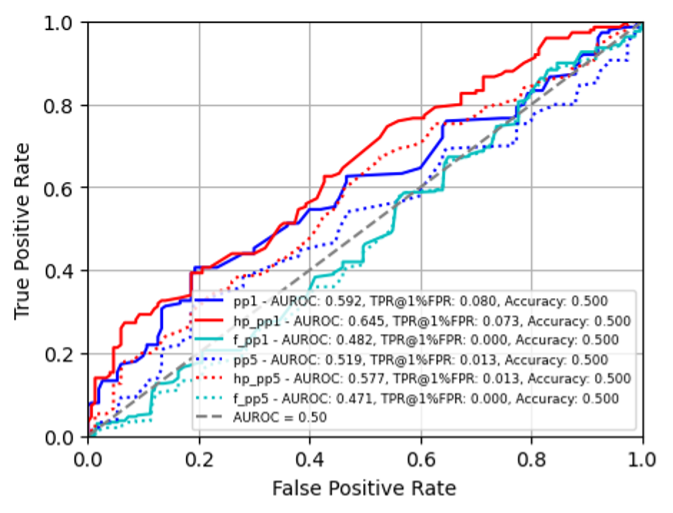}
    \includegraphics[scale=.32]{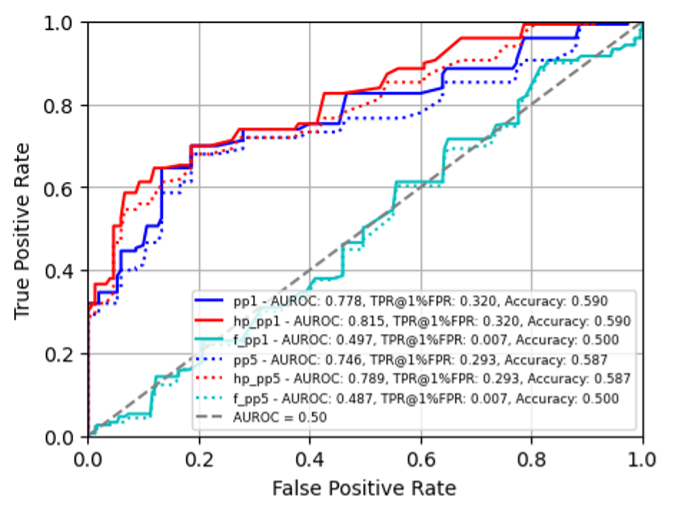}
\end{figure}

\end{document}